\documentclass[10pt,twocolumn,letterpaper]{article}

\usepackage{cvpr}              

\usepackage[dvipsnames]{xcolor}

\definecolor{cvprblue}{rgb}{0.21,0.49,0.74}
\usepackage[pagebackref,breaklinks,colorlinks,citecolor=cvprblue]{hyperref}

\usepackage{amsmath}
\usepackage{amssymb}
\usepackage[percent]{overpic}
\usepackage{cite}
\usepackage{verbatim}
\usepackage{bm}
\usepackage{multirow}
\usepackage{stfloats}
\usepackage{marvosym}

\usepackage{algorithmic}
\usepackage[linesnumbered, ruled]{algorithm2e}
\SetKwRepeat{Do}{do}{while}

\usepackage{colortbl}
\definecolor{Ocean}{RGB}{222,235,246}
\definecolor{orange}{RGB}{255,153,0}
\definecolor{blue1}{RGB}{144,170,220}

\usepackage[accsupp]{axessibility} 

\newcommand \footnoteONLYtext[1]
{
	\let \mybackup \thefootnote
	\let \thefootnote \relax
	\footnotetext{#1}
	\let \thefootnote \mybackup
	\let \mybackup \imareallyundefinedcommand
}

\begin{document}

\title{DiffMOT: A Real-time Diffusion-based Multiple Object Tracker with \\Non-linear Prediction\vspace{-10pt}}

\author{Weiyi Lv$^{1*}$\quad
Yuhang Huang$^{2*}$
\quad
Ning Zhang$^3$ \quad
Ruei-Sung Lin$^3$\quad
Mei Han$^3$\quad
Dan Zeng$^1$$^{\dag}$
\\
$^1$Shanghai University \quad $^2$National University of Defense Technology \quad
$^3$PAII Inc.\\
{\tt\small $^1$\texttt{\{kroery,dzeng\}@shu.edu.cn}, $^2$\texttt{huangai@nudt.edu.cn},}
{\tt\small $^3$\texttt{\{ning.zhang,rueisung,meihan\}@gmail.com}}\\
\href{https://diffmot.github.io/}{https://diffmot.github.io/}
}

\makeatletter
\def\thanks#1{\protected@xdef\@thanks{\@thanks
        \protect\footnotetext{#1}}}
\makeatother

\twocolumn[{%
\renewcommand\twocolumn[1][]{#1}%
\maketitle
\begin{center}
    \centering
    \captionsetup{type=figure}

\vspace{-15pt}
    \begin{overpic}
[width=1\linewidth,]{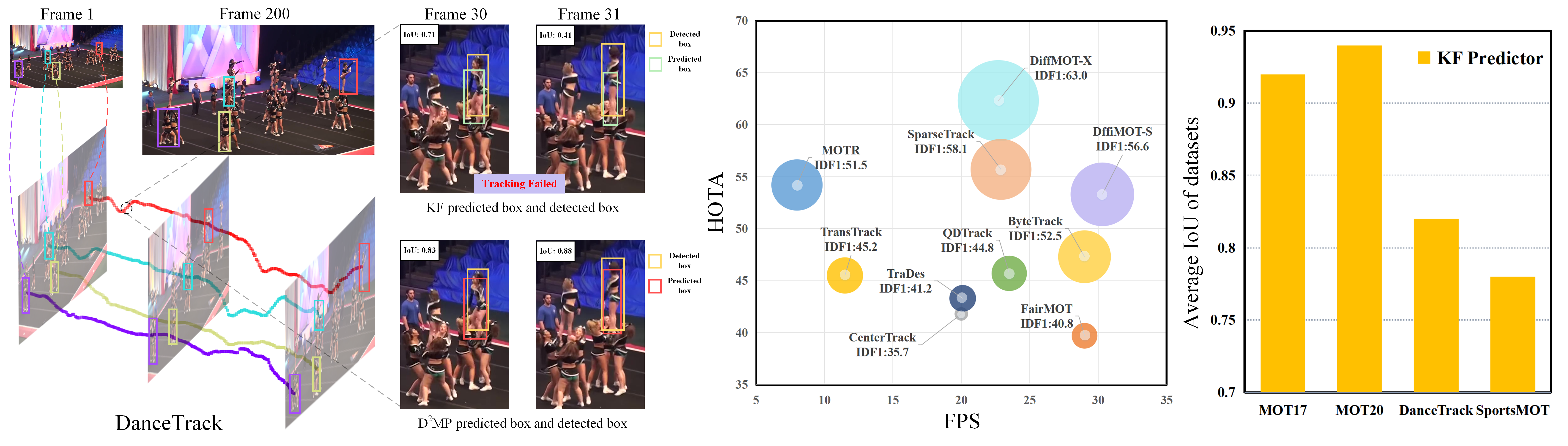}
\put(0,-0.5){\footnotesize{(a)} Tracklets visualization of DiffMOT and advance over KF tracker}
\put(52,-0.5){\footnotesize{(b)} Performance comparison}
\put(82,-0.5){\footnotesize{(c)} Dataset analysis}

    \end{overpic}
    \captionof{figure}
{
(a) illustrates the trajectories of DiffMOT on sampled sequences of DanceTrack. Each object's center position along the 200 frames is plotted in the 3D coordinates. The objects in DanceTrack exhibit non-linear motion trajectories.
Trackers with the KF predictor will fail in tracking in frame 30 for the inaccurate prediction, while our DiffMOT with D$^2$MP can track successfully.
(b) shows the HOTA-IDF1-FPS comparisons of different trackers. 
Our DiffMOT with the YOLOX-X detector achieves $62.3\%$ HOTA, $63.0\%$ IDF1 on the DanceTrack test set with $22.7$ FPS.
(c) shows the motion prediction of the linear Kalman Filter on different datasets.
The average IoU of the predicted and ground truth bounding boxes are used as the metric to demonstrate the linear (high IoU) and non-linear (low IoU) characteristics of each dataset.
}\label{fig1}
\end{center}%
}]

\begin{abstract}
In Multiple Object Tracking, objects often exhibit non-linear motion of acceleration and deceleration, with irregular direction changes. Tacking-by-detection (TBD) trackers with Kalman Filter motion prediction work well in pedestrian-dominant scenarios but fall short in complex situations when multiple objects perform non-linear and diverse motion simultaneously.
To tackle the complex non-linear motion, we propose a real-time diffusion-based MOT approach named DiffMOT. Specifically, for the motion predictor component, we propose a novel Decoupled Diffusion-based Motion Predictor (D$^2$MP). It models the entire distribution of various motion presented by the data as a whole. It also predicts an individual object's motion conditioning on an individual's historical motion information. Furthermore, it optimizes the diffusion process with much fewer sampling steps. As a MOT tracker, the DiffMOT is real-time at 22.7FPS, and also outperforms the state-of-the-art on DanceTrack\cite{(21)sun2022dancetrack} and SportsMOT\cite{(22)Cui_2023_ICCV} datasets with $62.3\%$ and $76.2\%$ in HOTA metrics, respectively. To the best of our knowledge, DiffMOT is the first to introduce a diffusion probabilistic model into the MOT to tackle non-linear motion prediction.
\footnoteONLYtext{$*$ equal contribution.}
\footnoteONLYtext{\dag corresponding author.}

\end{abstract}

\section{Introduction}
\label{sec1}
Multiple object tracking (MOT) is one of the fundamental computer vision tasks that aims at continuously tracking objects in video sequences. A successful MOT benefits downstream research such as action detection and recognition, pose tracking, and video understanding.
It also attracts much attention in various applications, including pedestrian-dominant smart-city, autonomous driving, and sports analysis.

Tracking-by-detection (TBD) paradigm has been a popular implementation in the MOT \cite{(9)zhang2022bytetrack, (11)cao2023observation, (30)du2023strongsort, (7)wojke2017simple, (8)bewley2016simple}. 
The TBD requires a robust detector and accurate motion predictor, which predicts an individual object's motion. 
These aforementioned approaches mainly focus on the pedestrian-dominant MOT17 \cite{(23)milan2016mot16} dataset whose objects possess linear motion patterns in terms of a constant-speed and mono-direction. 
In this way, the Kalman Filter (KF) is the natural choice because of its linear prediction and fast speed.

Other scenarios have objects moving \textit{non-linear motion} with less order and more variant. They are also less synchronous, which is different from the pedestrian-dominant scenes where people move more or less around the same speed. There are usually objects with (ac/de)celebration and irregular movement direction. 
Some typical \textit{non-linear motions} examples are dancers on the stage jumping all-directional \cite{(21)sun2022dancetrack}, or athletes in the field doing different movements \cite{(22)Cui_2023_ICCV}. In these cases,  motion predictors based on the constant-velocity assumption may not be accurate anymore. 
As depicted in Fig.~\ref{fig1} (c), linear KF fails to follow the detection ground truth on non-linear datasets, compared to its linear counterpart of the MOT17/20.

Some efforts \cite{(10)xiao2023motiontrack,(31)zhang2020multiple,(32)chaabane2021deft} attempt to tackle such \textit{non-linear motion} prediction using neural networks but not yet successful. 
This is due to either rigid network structures that are not flexible to obtain good adaptation, or heavy computation that is not suitable for the application.
For example, the vanilla neural network approaches (such as MLP and LSTM) \cite{(32)chaabane2021deft} can hardly get a satisfactory performance, while the optical flow-based and transformer-based methods \cite{(31)zhang2020multiple,(10)xiao2023motiontrack} have much lower FPS.
Naturally, our motivation is to devise an MOT tracker to achieve accurate non-linear motion prediction and real-time speed at the same time.

In this paper, we propose a novel multiple object tracker named DiffMOT with strong non-linear motion prediction and real-time speed. DiffMOT is based on the diffusion probabilistic model, which formulates bounding box position prediction as a denoising process conditioning on the previous bounding box motion trajectories. 
It offers two advantages over the traditional motion models with single-round learning and prediction. Specifically, at the learning stage, the multiple-step diffusion process has a thorough coverage of the input data with motion representations. During the prediction stage, an individual object's history motion is used as a condition to guide the denoising process for a better result.

Moreover, to improve the low efficiency of diffusion probabilistic models \cite{(34)ho2020denoising}, a Decoupled Diffusion-based Motion Predictor (D$^2$MP) optimizes the diffusion framework with both high efficiency and performance inspired by literature work \cite{(38)huang2023decoupled}.
Compared to the standard thousand-step sampling of a typical diffusion model, a one-step sampling process is devised with decoupled diffusion theory, reducing the inference time significantly.
As a result, DiffMOT achieves the best performances on two non-linear datasets (DanceTrack \cite{(21)sun2022dancetrack} and SportsMOT \cite{(22)Cui_2023_ICCV}) with real-time at $22.7$ FPS on an RTX 3090 machine. 

As we can see in Fig.~\ref{fig1} (b), our DiffMOT with a YOLOX-X detector combines both fast speed and superior performance, while methods with other motion predictors either suffer from low FPS (MOTR \cite{(6)zeng2022motr} and TransTrack \cite{(4)sun2020transtrack}) or low HOTA (FairMOT \cite{(1)zhang2021fairmot}, ByteTrack \cite{(9)zhang2022bytetrack}, and QDTrack \cite{(5)pang2021quasi}). 
Additionally, DiffMOT with a YOLOX-S detector can achieve much faster speed.

In summary, our contribution is three-fold:

\begin{itemize}

\item We propose a novel multiple object tracker named DiffMOT with strong non-linear motion prediction and real-time speed.
To the best of our knowledge, our work is the first to introduce a diffusion model into the MOT to tackle non-linear motion prediction.

\item We introduce a decoupled diffusion-based motion predictor D$^2$MP to model the motion distribution of objects with non-linear movements.
Compared to previous motion models, D$^2$MP excels in fitting the non-linear motion and fast inference speed.

\item DiffMOT outperforms SOTA methods on major public datasets in non-linear motions. DiffMOT achieves $62.3\%$ and $76.2\%$ HOTA metrics, on DanceTrack and SportsMOT, respectively. 

\end{itemize}

\section{Related Work}
\label{sec2}

\textbf{Motion Model in MOT}.
The motion model in MOT is employed to predict the future position of objects in the previous frame, categorized into linear and non-linear motion models.
The most classic linear motion model is the KF which is used in many literature, including the SORT \cite{(8)bewley2016simple}, DeepSORT \cite{(7)wojke2017simple}, FairMOT \cite{(1)zhang2021fairmot}, ByteTrack \cite{(9)zhang2022bytetrack}, OC-SORT \cite{(11)cao2023observation} and so on.
KF assumes that objects' motion velocity and direction remain constant within a small time interval.

For the non-linear motion models, certain literature proposes various non-linear approaches. For example, Optical flow-based motion models\cite{(31)zhang2020multiple} calculate pixel displacements between adjacent frames to obtain motion information, Long Short-Term Memory-based motion models\cite{(32)chaabane2021deft} capture sequence motion in latent space, and Transformer-based motion models\cite{(10)xiao2023motiontrack} capture long-range dependencies to model motion. 
However, none of the aforementioned can simultaneously achieve accurate non-linear motion prediction and real-time speed.
In this paper, the proposed D$^2$MP optimizes the typical diffusion framework and combines the superiority of performance and speed.

\textbf{Diffusion Probabilistic Models}.
Benefiting from the powerful fitting capabilities, DPMs have drawn extensive attention due to remarkable performances in image generation \cite{(47)dhariwal2021diffusion, (48)ho2022cascaded, (49)liu2023explicit, (50)park2022flexible, (51)zhao2021large}.
However, conventional DPMs suffer from a standard thousand-step sampling during inference, and recent efforts \cite{(38)huang2023decoupled,(40)song2020denoising,(41)lu2022dpm,(42)shirani2022thompson,(43)doucet2022score,(44)dockhorn2022genie} have focused on DPMs with few-step sampling.
DDM\cite{(38)huang2023decoupled} splits the typical diffusion process into two sub-processes to realize the few-step sampling. 
Inspired by DDM, we propose a specific decoupled diffusion-based motion predictor to perform motion prediction.
DiffusionTrack \cite{(39)luo2023diffusiontrack} is a concurrent work that utilizes diffusion models to construct the relationships between the paired boxes. In contrast, we focus on non-linear motion modeling and aim to learn the entire motion distribution.

\section{Method}
\label{sec3}

\subsection{Framework of DiffMOT}

\begin{figure}[t]
\centering
\begin{overpic}
[width=1\linewidth,]{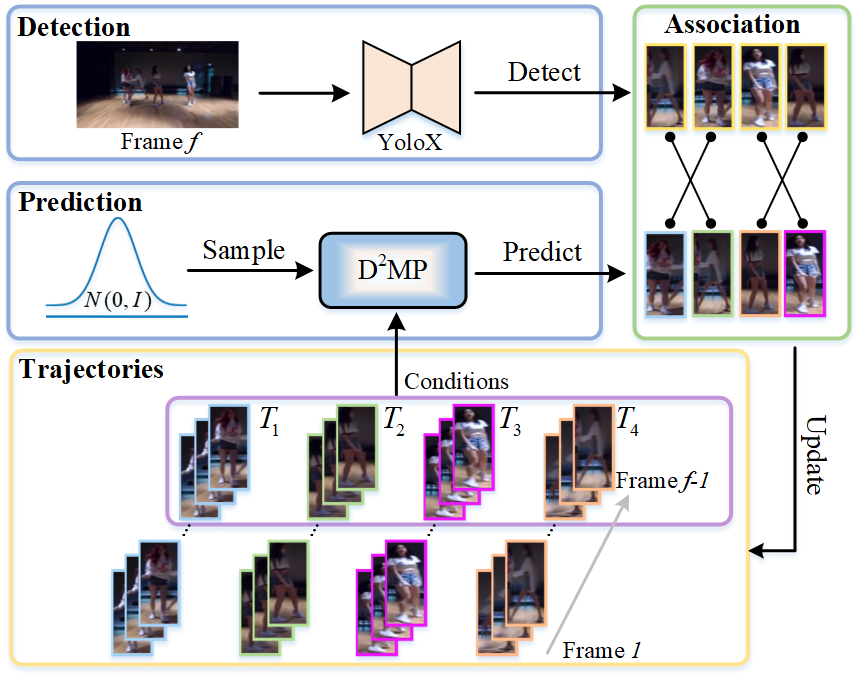}

\end{overpic}

\caption{The overall architecture of DiffMOT. DiffMOT consists of three parts: detection, motion prediction, and association.}
\vspace{-12pt}
\label{fig2}
\end{figure}

In this section, we introduce DiffMOT, a real-time diffusion-based MOT tracker, to track realistic objects with non-linear motion patterns. 
As shown in Fig.~\ref{fig2}, DiffMOT follows the tracking-by-detection framework that associates the detection of the current frame with the trajectories of the previous frame. 
The overall framework includes three parts: detection, motion prediction, and association.
Given a set of video sequences, DiffMOT first uses a detector to detect the objects' bounding boxes in the current frame.
Next, the future position of the target object in the previous frame is predicted via motion prediction.
The motion prediction is where our proposed D$^2$MP is devised. 
In particular, D$^2$MP is a diffusion-based motion predictor, which utilizes previous $n$ frames as conditions and generates the future motion of the objects from the previous frame.
Details of D$^2$MP are described in Sec.~\ref{sec:3.2}.
As a result, the motion prediction process will output the predicted bounding boxes of objects in the previous frame. 
Third and last, the association process matches the detected and predicted bounding boxes, thus updating the trajectories.

\textbf{Detection}.
We adopt the commonly used YoloX \cite{(28)ge2021yolox} as our detector. For a video sequence, the detector detects the bounding boxes of objects frame by frame. 

\textbf{Motion prediction}.
First, we retrieve the previous $n$ frames of information from the trajectories to serve as the condition for D$^2$MP.
Subsequently, we employ D$^2$MP to sample from the normal distribution, obtaining the motion of each object.
Finally, diffusion-based D$^2$MP generates the motion and finalizes the predicted bounding boxes of the current frame.

\textbf{Association}.
The association process is similar to the ByteTrack \cite{(9)zhang2022bytetrack}.
First, the predictions are matched with high-scoring bounding boxes from the detection using the Hungarian algorithm \cite{(52)kuhn1955hungarian}. The matching cost is defined by the re-id feature distance and  Intersection-over-Union(IoU). We also incorporate the dynamic appearance and adaptive weighting techniques as introduced in \cite{(14)maggiolino2023deep}.
Second, the unmatched predictions are matched with low-scoring boxes from detections using the Hungarian algorithm, employing the IoU as the cost function.
At last, we use the matched results to update the trajectories.

\subsection{Decoupled Diffusion-based Motion Predictor}\label{sec:3.2}
The decoupled diffusion-based motion predictor (D$^2$MP) aims to model the motion distribution of the entire dataset and treats motion prediction as a generative task. The algorithm generates future motion from a normal distribution conditioned on historical motion information.
We utilize the decoupled diffusion model \cite{(38)huang2023decoupled} to design the D$^2$MP. The decoupled diffusion model splits the typical data-to-noise process into two sub-processes: data to zero and zero to noise. The former decreases the clean data to zero gradually while the latter increases the zero data to the normal noise, and the summation of the two sub-processes makes up the data-to-noise process.
Incorporating the decoupled diffusion process into motion prediction, we carefully design the forward / reversed diffusion processes, network architecture, and training loss.

\begin{figure}[t]
\centering
\begin{overpic}
[width=1\linewidth,]{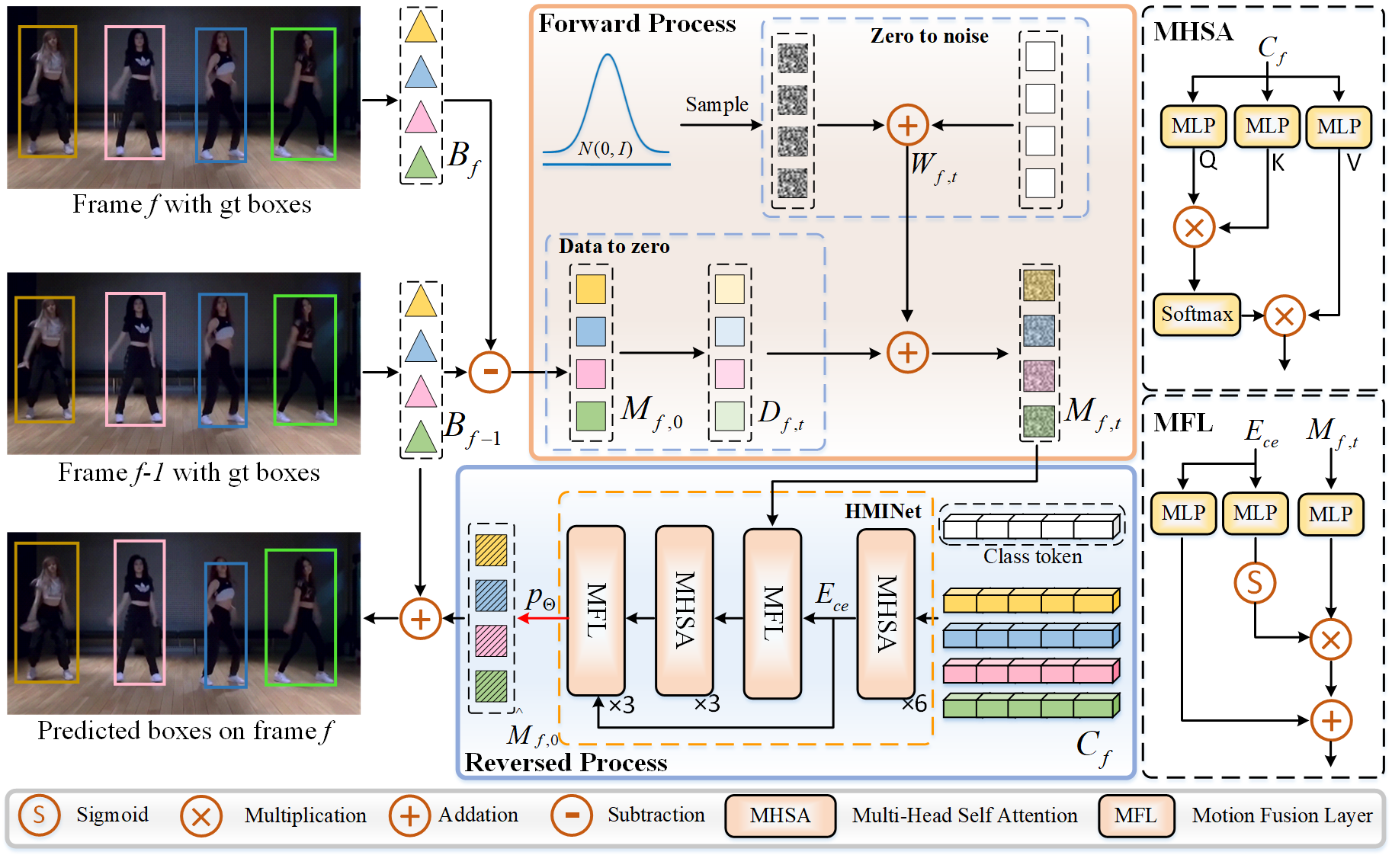}

\end{overpic}

\caption{The overall architecture of D$^2$MP. D$^2$MP consists of the forward process and the reversed process. In the forward process, data to zero and zero to noise processes are enclosed within the \textcolor{blue1}{blue} dashed box. In the reversed process, HMINet is enclosed within the \textcolor{orange}{orange} dashed box. $p_{\boldsymbol{\Theta}}$ refers to the operation introduced in Eq.~\ref{eq7} to reconstruct $\hat{\mathbf{M}}_{f, 0}$}.
\vspace{-15pt}
\label{fig3}
\end{figure}

D$^2$MP demonstrates a strong non-linear fitting capability with one-step sampling. The overall architecture of D$^2$MP is shown in Fig.~\ref{fig3}. Sec.~\ref{sec:3.2.1} describes the forward diffusion of adding noise to the motion data to a normal distribution. Sec.~\ref{sec:3.2.2} introduces the reversed diffusion to generate motion from a pure noise conditioned on historical motion information.
Sec.~\ref{sec:3.2.3} describes the proposed neural network used to parameterize the reversed process.

\subsubsection{Forward process}\label{sec:3.2.1}
In a sequence of MOT, all trajectories can be represented as $\textbf{Traj}=\{\mathbf{T}_1, \cdots, \mathbf{T}_{p}, \cdots, \mathbf{T}_{P}\}$, where $P$ denotes the number of trajectories.
Ignoring the subscript $p$, consider one of the object trajectories $\mathbf{T}=\{\mathbf{B}_{1}, \cdots, \mathbf{B}_{f}, \cdots, \mathbf{B}_{N}\}$, where $f$ is the frame index and $N$ is the total number of frames, $\mathbf{B}_{f}=(x_{f}, y_{f}, w_{f}, h_{f})$ is the object's bounding box representing the coordinates of the center point and the height and width of the box. We define the object motion at frame $f$ as the difference between current frame and previous frame:
\begin{equation}
\mathbf{M}_{f}=\mathbf{B}_{f}-\mathbf{B}_{f-1}=(\Delta x_{f}, \Delta y_{f}, \Delta w_{f}, \Delta h_{f}).
\label{eq4}
\end{equation}
In this paper, we define $\mathbf{M}_{f}$ as the clean motion data in the diffusion process.

Compared to the typical diffusion model \cite{(34)ho2020denoising} that only has a data-to-noise mapping, we follow the decoupled diffusion process \cite{(38)huang2023decoupled} that contains two sup-processes: data to zero and zero to noise.
To formulate the forward process, we introduce an additional subscript $t$, i.e. we use $\mathbf{M}_{f, 0}=\mathbf{M}_{f}$ and $\mathbf{M}_{f, t}$ to denote the clean and noisy motion data. Considering a continuous time axis $t\in [0, 1]$, we simultaneously conduct the data-to-zero and zero-to-noise processes over time $t$ to map the clean motion data $\mathbf{M}_{f, 0}$ to be a pure noise. Specifically, the data-to-zero process utilizes an analytic attenuation function to decrease the clean motion data to be zero data over time. In particular, we adopt the constant function as the analytic attenuation function and the data-to-zero process can be represented by:
\begin{equation}
	\mathcal{D}_{f, t} = \mathbf{M}_{f, 0}+ t\mathbf{c},
 \label{eq4-1}
\end{equation}
where $\mathbf{c}$ is the constant function and $\mathbf{c} = -\mathbf{M}_{f, 0}$ that can be obtained by solving $\mathbf{M}_{f, 0}+\int_{0}^{1} {\mathbf{c}\mathrm{d}t}=\mathbf{0}$ with reference to \cite{(38)huang2023decoupled}. 
In this way, the clean motion data is attenuated gradually over time $t$ and to be zero when $t=1$, i.e. $\mathcal{D}_{f, 1}=\mathbf{0}$.
At the same time, the zero-to-noise process adds the normal noise to the zero data gradually, increasing it to be the pure normal noise when $t=1$, which is written as:
\begin{equation}
	\mathcal{W}_{f, t} = \mathbf{0}+ \sqrt{t}\boldsymbol{z},
 \label{eq4-2}
\end{equation}
where $\boldsymbol{z}\sim \mathcal{N}(\mathbf{0}, \mathbf{I})$. Combining the two sub-processes, we can obtain the noisy motion data $\mathbf{M}_{f, t}$ through our forward process:
\begin{equation}
 \begin{aligned}
     \mathbf{M}_{f, t} &= \mathcal{D}_{f, t} + \mathcal{W}_{f, t}\\
                       &= \mathbf{M}_{f, 0} + t\mathbf{c} + \sqrt{t}\boldsymbol{z}.
 \end{aligned}
 \label{eq5}
\end{equation}

\subsubsection{Reversed process}\label{sec:3.2.2}
The reversed process utilizes the conditional probability $q(\mathbf{M}_{f, t-\Delta t}|\mathbf{M}_{f, t}, \mathbf{M}_{f, 0})$ to recover the object motion from the pure noise, $\mathbf{M}_{f, 1}$. The reconstructed motion is represented by $\hat{\mathbf{M}}_{f, 0}$.
Following \cite{(38)huang2023decoupled}, the reversed conditional probability $q(\mathbf{M}_{f, t-\Delta t}|\mathbf{M}_{f, t}, \mathbf{M}_{f, 0})$ can be written as:
\begin{equation}
\begin{aligned}
q(\mathbf{M}_{f, t-\Delta t}|&\mathbf{M}_{f, t}, \mathbf{M}_{f, 0})=\mathcal{N}(\mathbf{M}_{f, t-\Delta t}; \boldsymbol{\mu}, \boldsymbol{\Sigma})\\
&\boldsymbol{\mu}=\mathbf{M}_{f, t} - \Delta t\mathbf{c} -\frac{\Delta t}{\sqrt{t}}\boldsymbol{z}\\
&\boldsymbol{\Sigma}=\frac{\Delta t(t-\Delta t)}{t}\mathbf{I},
\end{aligned}
\label{eq6}
\end{equation}
where $\mathcal{N}$ represents the normal distribution, $\mathbf{\mu}$ and $\mathbf{\Sigma}$ are the mean and variance of $q(\mathbf{M}_{f, t-\Delta t}|\mathbf{M}_{f, t}, \mathbf{M}_{f, 0})$, and $\mathbf{I}$ is the identity matrix.
Inconveniently, $q(\mathbf{M}_{f, t-\Delta t}|\mathbf{M}_{f, t}, \mathbf{M}_{f, 0})$ cannot be accessed directly since $\boldsymbol{\mu}$ contains the unknown terms $\mathbf{c}$ and $\boldsymbol{z}$. Hence, we need to use the parameterized $p_{\boldsymbol{\Theta}}(\mathbf{M}_{f, t-\Delta t}|\mathbf{M}_{f, t})$ to approximate $q(\mathbf{M}_{f, t-\Delta t}|\mathbf{M}_{f, t}, \mathbf{M}_{f, 0})$.

\begin{equation}
\begin{aligned}
p_{\boldsymbol{\Theta}}(\mathbf{M}_{f, t-\Delta t}|&\mathbf{M}_{f, t})=\mathcal{N}(\mathbf{M}_{f, t-\Delta t}; \boldsymbol{\mu}_{\boldsymbol{\Theta}},
\boldsymbol{\Sigma})\\
\boldsymbol{\mu}_{\boldsymbol{\Theta}} = \mathbf{M}_{f, t} - \Delta t\mathbf{c}_{\boldsymbol{\Theta}}(&\mathbf{M}_{f, t}, t, \mathbf{C}_{f})-\frac{\Delta t}{\sqrt{t}}\boldsymbol{z}_{\boldsymbol{\Theta}}(\mathbf{M}_{f, t}, t, \mathbf{C}_{f})\\
&\boldsymbol{\Sigma}=\frac{\Delta t(t-\Delta t)}{t}\mathbf{I}.
\end{aligned}
\label{eq7}
\end{equation}
Here, $\mathbf{c}_{\boldsymbol{\Theta}}(\mathbf{M}_{f, t}, t, \mathbf{C}_{f})$ and $\boldsymbol{z}_{\boldsymbol{\Theta}}(\mathbf{M}_{f, t}, t, \mathbf{C}_{f})$ are parameterized via a neural network $\boldsymbol{\Theta}$.
Additionally, $\mathbf{C}_{f}$ is the conditioned historical information whose details will be described in Sec.~\ref{sec:3.2.3}.
For the simplicity, we replace $\mathbf{c}_{\boldsymbol{\Theta}}(\mathbf{M}_{f, t}, t, \mathbf{C}_{f})$ and $\boldsymbol{z}_{\boldsymbol{\Theta}}(\mathbf{M}_{f, t}, t, \mathbf{C}_{f})$ with $\mathbf{c}_{\boldsymbol{\Theta}}$ and $\boldsymbol{z}_{\boldsymbol{\Theta}}$ in the following paper.

The original decoupled diffusion model proposes a two-branch architecture to parameterize $\mathbf{c}$ and $\boldsymbol{z}$ simultaneously, however, it consumes additional computational cost, hurting its efficiency. 
To speed up the process, we propose a specific reversed process that only needs to parameterize $\mathbf{c}$ and maintain the performance.
From Eq.~\ref{eq5}, we can represent $\boldsymbol{z}$ by:
\begin{equation}
 \begin{aligned}
\boldsymbol{z} &= \frac{1}{\sqrt{t}}(\mathbf{M}_{f, t}-\mathbf{M}_{f, 0}-t\mathbf{c})\\
&=\frac{1}{\sqrt{t}}(\mathbf{M}_{f, t}-(t-1)\mathbf{c}).
\end{aligned}
 \label{eq8}
\end{equation}
In the similar way, we can represent $\boldsymbol{z}_{\Theta}$ by $\mathbf{M}_{f, t}$ and $\mathbf{c}_{\Theta}$:
\begin{equation}
\boldsymbol{z}_{\boldsymbol{\Theta}} =\frac{1}{\sqrt{t}}(\mathbf{M}_{f, t}-(t-1)\mathbf{c}_{\boldsymbol{\Theta}}).
 \label{eq9}
\end{equation}
Substituting Eq.~\ref{eq9} into Eq.~\ref{eq7}, we have:
\begin{equation}
    \boldsymbol{\mu}_{\boldsymbol{\Theta}} = \frac{t-\Delta t}{t}\mathbf{M}_{f, t} - \frac{\Delta t}{t}\mathbf{c}_{\boldsymbol{\Theta}}
    \label{eq10}
\end{equation}
Therefore, we only need $\mathbf{c}_{\boldsymbol{\Theta}}$ to solve the reversed process.
In general, solving the reversed process of typical diffusion models \cite{(34)ho2020denoising} needs thousands of steps via numerical integration. Differently, benefiting from the analyticity of the decoupled diffusion process \cite{(38)huang2023decoupled}, the proposed reversed process enables one-step sampling, which further increases the inference speed to achieve real-time performance.

\subsubsection{Historical Memory Information Network (HMINet)}\label{sec:3.2.3}
In this section, we introduce a specific historical memory information network to parameterize $\mathbf{c}_{\boldsymbol{\Theta}}$ in Eq.~\ref{eq10}.
HMINet first utilizes the multi-head self-attention layers to extract the condition embedding from the conditioned input $\mathbf{C}_{f}$. Afterward, we use the condition embedding as guidance and integrate it into the noisy motion feature. Finally, we use an MLP layer to obtain the final prediction $\mathbf{c}_{\boldsymbol{\Theta}}$.

\textbf{Extracting condition embedding}.
As the literature shows \cite{(45)rombach2022high, (46)zhang2023adding}, the conditioned guidance is important in providing clues for generating the final result. Figure \ref{fig3} illustrates the details.
We leverage the motion information of previous $n$ frames as conditioned input $\mathbf{C}_{f}$.
The motion information of frame $f$ is defined as the combination of the object's bounding box and movement:
\begin{equation}
I_{f}=(x_{f}, y_{f}, w_{f}, h_{f}, \Delta x_{f}, \Delta y_{f}, \Delta w_{f}, \Delta h_{f}).
\label{eq11}
\end{equation}
Thus, the conditioned input is represented by: $\mathbf{C}_{f}=[I_{f-1}; I_{f-2}, ..., I_{f-n}], \mathbf{C}_{f}\in\mathbb{R}^{n\times 8}$.
We aim to capture the long-range dependencies between different frames in $\mathbf{C}_{f}$, so we use the multi-head self-attention (MHSA) as the base layer.
In practice, we feed $\mathbf{C}_{f}$ to $6\times$ MHSA layers to extract condition embedding.
We concatenate a learnable \emph{class token} $\mathbf{E}\in\mathbb{R}^{1\times 512}$ and $\mathbf{C}_{f}$, and then incorporate historical motion information contained in $\mathbf{C}_{f}$ into the \emph{class token} using multi-head self-attention layers. In this way, we use the updated \emph{class token} as condition embedding, denoted by $\mathbf{E}_{ce}\in\mathbb{R}^{1\times 512}$.

\textbf{Incorporating $\mathbf{E}_{ce}$ into noisy motion feature}.
After obtaining the condition embedding $E_{ce}$, we aim to incorporate its information into the noisy motion data. In particular, we construct a motion fusion layer (MFL) that fuses $\mathbf{E}_{ce}$ and $\mathbf{M}_{f, t}$ into a unified motion feature $\overline{\mathbf{M}}_{f, t}$. Specifically, MFL utilizes an MLP layer to encode $\mathbf{M}_{f, t}$, letting the dimension of $\mathbf{M}_{f, t}$ be same as $\mathbf{E}_{ce}$. At the same time, the other two MLP layers are applied to $\mathbf{E}_{ce}$, generating two variables that are used as the scale and shift coefficients. The above process can be described by:
\begin{equation}
    \overline{\mathbf{M}}_{f, t} =\operatorname{Sigmoid}(\operatorname{MLP}(\mathbf{E}_{ce}))\cdot\operatorname{MLP}(\mathbf{M}_{f, t})
    +\operatorname{MLP}(\mathbf{E}_{ce}).
\label{E12}
\end{equation}
Afterward, we concatenate $\mathbf{E}_{ce}$ and $\overline{\mathbf{M}}_{f, t}$ together and feed it into the stacked MHSA and MFL layers to conduct the further feature fusion. Finally, the fused feature is fed into an MLP layer to get the final prediction $\mathbf{c}_{\boldsymbol{\Theta}}$.
After obtaining $\mathbf{c}_{\boldsymbol{\Theta}}$, we can use Eq.~\ref{eq7} to reconstruct $\hat{\mathbf{M}}_{f, 0}$.

\subsubsection{Training loss}
We use the ground truth $\mathbf{c}$ to supervise the network to optimize the $\boldsymbol{\Theta}$.
We adopt the smooth L1 loss and the final loss function is written as:
\begin{equation}
L = 
\left\{
\begin{aligned}
0.5 (\mathbf{c}_{\boldsymbol{\Theta}}- \mathbf{c})^2 & &  |\mathbf{c}_{\boldsymbol{\Theta}}- \mathbf{c}|< 1&\\
|\mathbf{c}_{\boldsymbol{\Theta}}- \mathbf{c}|-0.5 & & otherwise.&
\end{aligned}
\right.
\label{eq13}
\end{equation}

\section{Experiments}
\label{sec4}

\begin{table}[t]
\begin{center}

\begin{tabular}{l |  c   c   c   c   c}
\toprule

\makebox[0.03\textwidth][l]{Method}  & \makebox[0.04\textwidth][c]{HOTA$\uparrow$} & \makebox[0.03\textwidth][c]{IDF1$\uparrow$} & \makebox[0.03\textwidth][c]{AssA$\uparrow$} & \makebox[0.03\textwidth][c]{MOTA$\uparrow$} & \makebox[0.03\textwidth][c]{DetA$\uparrow$} \\

\midrule
FairMOT\cite{(1)zhang2021fairmot} & 39.7 & 40.8 & 23.8 & 82.2 & 66.7\\
CenterTrack\cite{(2)zhou2020tracking} & 41.8 & 35.7 & 22.6 & 86.8 & 78.1\\
TraDes\cite{(3)wu2021track}     & 43.3 & 41.2 & 25.4 & 86.2 & 74.5\\
TransTrack\cite{(4)sun2020transtrack}   & 45.5 & 45.2 & 27.5 & 88.4 & 75.9\\
QDTrack\cite{(5)pang2021quasi} & 45.7 & 44.8 & 29.2 & 83.0 & 72.1\\
\small{DiffusionTrack\cite{(39)luo2023diffusiontrack}} & 52.4 & 47.5 & 33.5 & 89.5 & 82.2\\
MOTR\cite{(6)zeng2022motr}     & 54.2 & 51.5 & 40.2 & 79.7 & 73.5\\
\rowcolor{Ocean} DeepSORT\cite{(7)wojke2017simple} & 45.6 & 47.9 & 29.7 & 87.8 & 71.0\\
\rowcolor{Ocean} ByteTrack\cite{(9)zhang2022bytetrack} & 47.3 & 52.5 & 31.4 & 89.5 & 71.6\\
\rowcolor{Ocean} SORT\cite{(8)bewley2016simple}     & 47.9 & 50.8 & 31.2 & 91.8 & 72.0\\
\rowcolor{Ocean} MotionTrack\cite{(10)xiao2023motiontrack} & 52.9 & 53.8 & 34.7 & 91.3 & 80.9\\
\rowcolor{Ocean} OC-SORT\cite{(11)cao2023observation}  & 55.1 & 54.2 & 38.0 & 89.4 & 80.3\\
\rowcolor{Ocean} StrongSORT\cite{(30)du2023strongsort}  & 55.6 & 55.2 & 38.6 & 91.1 & 80.7\\
\rowcolor{Ocean} SparseTrack\cite{(12)liu2023sparsetrack}  & 55.7 & 58.1 & 39.3 & 91.3 & 79.2\\
\rowcolor{Ocean} C-BIoU\cite{(13)yang2023hard}  & 60.6 & 61.6 & 45.4 & 91.6 & 81.3\\
\rowcolor{Ocean} \small{Deep OC-SORT\cite{(14)maggiolino2023deep}}  & 61.3 & 61.5 & 45.8 & 92.3 & 82.2\\
\rowcolor{Ocean} DiffMOT & \bf 62.3 & \bf 63.0 & \bf 47.2 & \bf 92.8 & \bf 82.5\\
\bottomrule
\end{tabular}
\caption{ 
Comparison with SOTA MOT trackers on the DanceTrack test sets without using any extra training data. 
Trackers in the blue block use the same YOLOX detector. 
$\uparrow$ means the higher the better and $\downarrow$ means the lower the better. \textbf{Bold} numbers indicate the best result. 
}
\vspace{-20pt}
\label{T1}

\end{center}
\end{table}

\begin{table}[t]
\begin{center}

\begin{tabular}{ l |  c    c  c   c   c   c}
\toprule
\makebox[0.03\textwidth][l]{Method}  & \makebox[0.04\textwidth][c]{HOTA$\uparrow$} & \makebox[0.03\textwidth][c]{IDF1$\uparrow$} & \makebox[0.03\textwidth][c]{AssA$\uparrow$} & \makebox[0.03\textwidth][c]{MOTA$\uparrow$} & \makebox[0.03\textwidth][c]{DetA$\uparrow$} \\

\midrule

FairMOT\cite{(1)zhang2021fairmot} & 49.3 & 53.5 & 34.7 & 86.4 & 70.2\\
GTR\cite{(15)zhou2022global} & 54.5 & 55.8 & 45.9 & 67.9 & 64.8\\
QDTrack\cite{(5)pang2021quasi} & 60.4 & 62.3 & 47.2 & 90.1 & 77.5\\
CenterTrack\cite{(2)zhou2020tracking} & 62.7 & 60.0 & 48.0 & 90.8 & 82.1\\
TransTrack\cite{(4)sun2020transtrack} & 68.9 & 71.5 & 57.5 & 92.6 & 82.7\\
\rowcolor{Ocean} ByteTrack\cite{(9)zhang2022bytetrack} & 62.8 & 69.8 & 51.2 & 94.1 & 77.1\\
\rowcolor{Ocean} BoT-SORT\cite{(16)aharon2022bot} & 68.7 & 70.0 & 55.9 & 94.5 & 84.4\\
\rowcolor{Ocean} OC-SORT\cite{(11)cao2023observation} & 71.9 & 72.2 & 59.8 & 94.5 &  86.4\\
\rowcolor{Ocean} DiffMOT &  72.1 &  72.8 &  60.5 &  94.5 &  86.0\\
\midrule
\rowcolor{Ocean} *ByteTrack\cite{(9)zhang2022bytetrack} & 64.1 & 71.4 & 52.3 & 95.9 & 78.5\\
\rowcolor{Ocean} *MixSort-Byte\cite{(22)Cui_2023_ICCV} &  65.7 &  74.1 &  54.8 &  96.2 & 78.8\\
\rowcolor{Ocean} *OC-SORT\cite{(11)cao2023observation} & 73.7 & 74.0 & 61.5 & 96.5 &  88.5\\
\rowcolor{Ocean} *MixSort-OC\cite{(22)Cui_2023_ICCV} &  74.1 &  74.4 &  62.0 &  96.5 & 88.5\\
\rowcolor{Ocean} *DiffMOT & \bf 76.2 & \bf 76.1 & \bf 65.1 & \bf 97.1 & \bf 89.3\\

\midrule
\end{tabular}
\caption{ 
Comparison with SOTA MOT trackers on the SportsMOT test sets. 
Following the MixSORT \cite{(22)Cui_2023_ICCV} convention, the methods with * indicate that their detectors are trained on the SportsMOT train and validation sets.
}
\vspace{-20pt}
\label{T2}
\end{center}
\end{table}

\subsection{Datasets and Evaluation Metrics}
\textbf{Datasets.} 
We conducted the main experiments on DanceTrack \cite{(21)sun2022dancetrack} and SportsMOT \cite{(22)Cui_2023_ICCV} datasets
in which the objects possess non-linear motion patterns. 
DanceTrack is a dataset comprised of dance videos, consisting of 40 training sequences, 25 validation sequences, and 35 testing sequences. 
DanceTrack exhibits a highly similar appearance with complex non-linear motion patterns such as irregular direction changes.
Therefore, DanceTrack's evaluation requires a significant demand on the trackers' capacity to robustly handle non-linear motions.
SportsMOT introduces video sequences from three different sporting events: soccer, basketball, and volleyball.
The dataset comprises a total of 45 training sequences, 45 validation sequences, and 150 testing sequences.
It displays extensive acceleration and deceleration motions, thereby demanding robustness in handling non-linear motions from trackers.
MOT17 \cite{(23)milan2016mot16} is a conventional and commonly used pedestrian-dominant dataset in MOT. The motion patterns of objects in MOT17 are approximated linearly.

\textbf{Metrics}.
We utilize Higher Order Metric \cite{(26)luiten2021hota} (HOTA, AssA, DetA), IDF1 \cite{(27)ristani2016performance}, and CLEAR metrics \cite{(25)bernardin2008evaluating} (MOTA) as our evaluation metrics.
Among various metrics, HOTA is the primary metric that explicitly balances the effects of performing accurate detection and association.
IDF1 and AssA are used for association performance evaluation.
DetA and MOTA primarily evaluate detection performance.
Furthermore, we employ the frames per second (FPS) metric to assess the speed of the algorithm.

\textbf{Implementation Details}.
In the training stage, we set the smallest time step of the diffusion process to $0.001$. 
We select previous $n=5$ frames of historical motion information as the condition.
For the optimization, we adopt Adam optimizer training for $800$ epochs. The learning rate is set to $10^{-4}$. The batch size is set to $2048$.
All experiments are trained on 4 GeForce RTX 3090 GPUs.
In experiments, we use the YOLOX-X as the default detector as recent works\cite{(16)aharon2022bot, (11)cao2023observation} for a fair comparison unless there is a specific announcement.
For the denoising process, we employ D$^2$MP to conduct one-step sampling from a standard normal distribution to obtain the objects' motion.
In the association stage, similar to ByteTrack, we perform separate matching for high-scoring and low-scoring boxes with the high threshold $\tau_{high}=0.6$ and the low threshold $\tau_{low}=0.4$.
Besides, we adopt the same dynamic appearance and adaptive weighting techniques as \cite{(16)aharon2022bot}.

\begin{table}[t]
\begin{center}

\begin{tabular}{l |  c   c   c   c   c}
\toprule
\makebox[0.03\textwidth][l]{Method}  & \makebox[0.04\textwidth][c]{HOTA$\uparrow$} & \makebox[0.03\textwidth][c]{IDF1$\uparrow$} & \makebox[0.03\textwidth][c]{AssA$\uparrow$} & \makebox[0.03\textwidth][c]{MOTA$\uparrow$} & \makebox[0.03\textwidth][c]{DetA$\uparrow$} \\

\midrule

\small{DiffusionTrack\cite{(39)luo2023diffusiontrack}} & 60.8 & 73.8 & 58.8 & 77.9 & 63.2\\
 MotionTrack\cite{(10)xiao2023motiontrack} & 61.6 & 75.1 & 60.2 & 78.6 & -\\
 ByteTrack\cite{(9)zhang2022bytetrack} & 63.1 & 77.3 & 62.0 &  80.3 &  64.5\\
 OC-SORT\cite{(11)cao2023observation} &  63.2 &  77.5 &  63.4 & 78.0 & 63.2\\ 
 MixSort-OC\cite{(22)Cui_2023_ICCV} & 63.4 &  77.8 &  63.2 &  78.9 & 63.8\\
 MixSort-Byte\cite{(22)Cui_2023_ICCV} & 64.0 &  78.7 &  64.2 &  79.3 & 64.1\\
 C-BIoU\cite{(13)yang2023hard} &  64.1 &  79.7 &  63.7 & 79.7 & 64.8\\ 
 StrongSORT\cite{(30)du2023strongsort} &  64.4 &  79.5 &  64.4 & 79.6 & 64.6\\
 \small{Deep OC-SORT\cite{(14)maggiolino2023deep}}  & 64.9 & \bf 80.6 & \bf 65.9 & 79.4 & 64.1\\
 SparseTrack\cite{(12)liu2023sparsetrack}  & \bf 65.1 & 80.1 & 65.1 & \bf 81.0 & \bf 65.3\\
 DiffMOT & 64.5 & 79.3 & 64.6 & 79.8 & 64.7\\
\bottomrule
\end{tabular}
\caption{ 
Comparison with SOTA MOT trackers on the MOT17 test sets under the “private detector” protocol. 
All methods use the same YOLOX detector.
}
\label{T3}
\vspace{-20pt}
\end{center}
\end{table}

\subsection{Benchmark Evaluation}
We conduct the experiments on the test set of DanceTrack and SportsMOT with non-linear motion patterns to demonstrate the effectiveness of our model. We put all methods that use the same YOLOX detector results in the blue block. 
We also compare the tracking performances on the test set of MOT17 under the "private detection" protocol to demonstrate that our tracker can achieve satisfactory performance even in pedestrian-dominant scenarios.

\textbf{DanceTrack.}
We report DiffMOT's performance on DanceTrack in Tab.~\ref{T1}.
It can be seen that DiffMOT consistently achieves the best results across all metrics with the $62.3\%$ HOTA, $63.0\%$ IDF1, $47.2\%$ AssA, $92.8\%$ MOTA, and $82.5\%$ DetA.
Compared with the previous SOTA tracker Deep OC-SORT, DiffMOT outperforms it in HOTA by $1.0\%$.
Moreover, we would like to emphasize the improvement of the association-related metrics, such as IDF1 and AssA.
DiffMOT outperforms the Deep OC-SORT in IDF1 and AssA by $1.5\%$ and $1.4\%$, respectively.
The results demonstrate the robustness of the proposed DiffMOT in dealing with rich non-linear motion.
Note that DiffusionTrack is a concurrent work similar to ours that introduces the diffusion model in MOT. 
They employ the DDPM\cite{(34)ho2020denoising} to model the distribution of the relationship between paired boxes without considering the non-linear motion of objects. 
Differently, we model the entire motion distribution and generate future motion from a normal distribution during the prediction process.
In comparison to DiffusionTrack, we make full use of the diffusion model's strong fitting capabilities and we can see in the table that DiffMOT outperforms it in HOTA by $8.1\%$.

Additionally, we conduct experiments on the test sets of Dancetrack using different detectors as shown in Tab.~\ref{T9}.
We observe that more refinements cost brings more performance gain and results in less FPS. 
We observed that smaller detectors can achieve higher FPS. DiffMOT can achieve the $30.3$ FPS with the YOLOX-S detector.
This indicates that DiffMOT can flexibly choose different detectors for various real-world application scenarios.

\begin{table}[t]
\begin{center}

\begin{tabular}{l |  c   c   c   c}
\toprule

Detector  & HOTA$\uparrow$ & IDF1$\uparrow$ & MOTA$\uparrow$ & FPS$\uparrow$ \\
\midrule
YOLOX-S & 53.3 & 56.6 & 88.4 & \bf 30.3\\
YOLOX-M & 57.2 & 58.6 & 91.2 & 25.4\\
YOLOX-L & 61.5 & 61.7 & 92.0 & 24.2\\
YOLOX-X & \bf 62.3 & \bf 63.0 & \bf 92.8 & 22.7\\
\bottomrule
\end{tabular}
\caption{ 
Comparison of different detectors in DiffMOT on the DanceTrack test sets. The best results are shown in bold.
}
\label{T9}
\vspace{-20pt}
\end{center}
\end{table}

\textbf{SportsMOT.}
To further demonstrate the performance of DiffMOT in other non-linear scenarios, we conduct experiments on the SportsMOT benchmark which is characterized by a large amount of (ac/de)celeration object motion.
Following \cite{(22)Cui_2023_ICCV}, we conduct experiments under two different detector setups.
As shown in Tab.~\ref{T2}, methods with * indicate their detectors are trained on both train and validation sets, while others are trained only on the train set.
DiffMOT achieves SOTA results in both setups.
In the train-only setup, DiffMOT obtains $72.1\%$ HOTA, $72.8\%$ IDF1, $60.5\%$ AssA, $94.5\%$ MOTA, and $86.0\%$ DetA, which surpasses previous methods comprehensively.
In the other setup, DiffMOT also achieves the best performance across all metrics and outperforms the previous SOTA tracker MixSort-OC by $2.1\%$ in HOTA, $1.7\%$ in IDF1, $3.1\%$ in AssA, $0.6\%$ in MOTA, and $0.8\%$ in DetA with the $76.2\%$ HOTA, $76.1\%$ IDF1, $65.1\%$ AssA, $97.1\%$ MOTA, and $89.3\%$ DetA.
The results further indicate that DiffMOT exhibits strong robustness in scenarios with (ac/de)celeration object motion.

\textbf{MOT17.}
We also conduct the experiment on the conventional and commonly used pedestrian-dominant MOT17 dataset.
On MOT17, it can be seen from the Tab.~\ref{T3} that DiffMOT achieves $64.5\%$ HOTA, $79.3\%$ IDF1, $64.6\%$ AssA, $79.8\%$ MOTA, and $64.7\%$ DetA.
Although the proposed DiffMOT is designed specifically for non-linear motion scenes, it can still achieve comparable performances to other SOTA methods under pedestrian-dominant scenarios.

\begin{table}[t]
\begin{center}

\begin{tabular}{l |  c   c   c   c   c}
\toprule
\makebox[0.06\textwidth][l]{Method}  & \makebox[0.04\textwidth][c]{HOTA$\uparrow$} & \makebox[0.04\textwidth][c]{IDF1$\uparrow$} & \makebox[0.04\textwidth][c]{AssA$\uparrow$} & \makebox[0.04\textwidth][c]{MOTA$\uparrow$} & \makebox[0.04\textwidth][c]{DetA$\uparrow$} \\

\midrule
IoU Only & 44.7 & 36.8 & 25.3 & 87.3 & \bf 79.6\\
Kalman Filter & 46.8 & 52.1 & 31.3 & 87.5 & 70.2\\
LSTM             & 51.2 & 51.6 & 34.3 & 87.1 & 76.7\\
Transformer      & 54.6 & 54.6 & 38.1 & 89.2 & 78.6\\
D$^2$MP(ours)      & \bf 55.7 & \bf 55.2 & \bf 39.5 & \bf 89.3 & 78.9\\
\bottomrule
\end{tabular}
\caption{ 
Comparison of different motion models on the DanceTrack validation sets. The best results are shown in bold.
}
\label{T5}
\vspace{-10pt}
\end{center}
\end{table}

\begin{table}[t]
\begin{center}

\begin{tabular}{l  p{0.0755\textwidth} |  c   c   c  c}
\toprule
\makebox[0.03\textwidth][l]{Method}  & sampling steps & \makebox[0.03\textwidth][c]{HOTA$\uparrow$} & \makebox[0.03\textwidth][c]{IDF1$\uparrow$} & \makebox[0.03\textwidth][c]{MOTA$\uparrow$} & \makebox[0.03\textwidth][c]{FPS$\uparrow$}\\

\noalign{\smallskip}
\midrule
D$^2$MP-TB & 1 & 44.5 & 48.0 & 89.2 & 20.9\\
D$^2$MP-TB & 10 & 52.3 & 50.5 &  89.2 & 13.1\\
D$^2$MP-TB & 20 & 54.6 & 53.6 &  \bf 89.3 & 7.5\\
D$^2$MP-OB   & 1  & \bf 55.7 & \bf 55.2 &  \bf 89.3 & \bf 22.7\\
\bottomrule
\end{tabular}
\caption{  
Comparison of D$^2$MP with two or one branch on the DanceTrack validation sets. 
The best results are shown in bold.
}
\label{T10}
\vspace{-20pt}
\end{center}
\end{table}

\subsection{Ablation Studies}
We conduct ablation studies on the validation set of DanceTrack.
we would like to note that all ablation study focuses on the motion prediction step based on the D$^2$MP. We use the same  YOLOX-X detector throughout all experiments.
The ablation studies focus on investigating the impact of different motion models, different architectures of D$^2$MP, different conditions, and the length of historical motion information on the proposed DiffMOT.

\textbf{Different motion models.}
To compare our method with other motion models regarding its modeling capability in non-linear motion scenarios, we conduct the experiment in Tab.~\ref{T5}.
The motion models we compared include the linear motion model (KF) and non-linear motion models (LSTM and Transformer-based motion models).
We also present IoU only association, without any motion model for reference.
From the table, it can be seen that our motion model achieves the best performance with $55.7\%$ in HOTA, $55.2\%$ in IDF1, $39.5\%$ in AssA, $89.3\%$ in MOTA, and $78.9\%$ in DetA.
This demonstrates that D$^2$MP can ensure more robust non-linear motion predictions because our diffusion-based motion model directly learns the distribution of all objects motion across the entire dataset rather than individual object trajectories.

\textbf{Different architectures of D$^2$MP.}
We compared the experimental results of different D$^2$MP architectures which are mentioned in Sec.~\ref{sec:3.2.2} in Tab.~\ref{T10}.
In the table, "D$^2$MP-TB" refers to the two-branch architecture, which optimizes both $\mathbf{c}_{\boldsymbol{\Theta}}$ and $\mathbf{z}_{\boldsymbol{\Theta}}$ simultaneously.
During the reversed process, D$^2$MP-TB uses the Eq.~\ref{eq7} to approximate $q(\mathbf{M}_{f, t-\Delta t}|\mathbf{M}_{f, t}, \mathbf{M}_{f, 0})$.
On the other hand, "D$^2$MP-OB" refers to the one-branch architecture that is used in our DiffMOT. 
The one-branch architecture only requires the optimization of $\mathbf{c}_{\boldsymbol{\Theta}}$, and during the reversed process, D$^2$MP-OB uses the Eq.~\ref{eq9} 
to approximate $q(\mathbf{M}_{f, t-\Delta t}|\mathbf{M}_{f, t}, \mathbf{M}_{f, 0})$.
As shown in the table, D$^2$MP-OB outperforms D$^2$MP-TB by $11.2\%$ in HOTA when using one-step sampling. 
This is attributed to the increased complexity of learning in the two-branch network.
To achieve comparable performance to the D$^2$MP-OB, the D$^2$MP-TB requires multiple sampling steps.
D$^2$MP-TB can achieve $54.6\%$ HOTA, $53.6\%$ IDF1, and $89.3\%$ MOTA after 20 sampling steps. 
However, at this point, the speed is only $7.5$ FPS, making it hard for practical applications.

\textbf{Different conditions.}
To demonstrate the advantages of our designed condition used in HMINet, we conducted the ablation experiment as shown in Tab.~\ref{T7}.
In the table, $\mathbf{B}_{f-1}$ refers to using the object box from the previous frame as a condition which is defined as $\mathbf{B}_{f-1}=(x_{f-1}, y_{f-1}, w_{f-1}, h_{f-1})$.
$\mathbf{M}_{f-1}$ refers to the object motion which is defined as Eq.~\ref{eq4}.
$\mathbf{I}_{f-1}$ refers to the motion information which is used in our DiffMOT. It is defined as Eq.~\ref{eq11}.
From the table, it can be seen that utilizing $\mathbf{I}_{f-1}$ as the condition yields the best results.
This is because $\mathbf{M}_{f-1}$ contains only the positional information of the bounding boxes, using it as a condition can lead to motion predictions with larger deviations.
On the other hand, $\mathbf{I}_{f-1}$ contains only the information on the motion change, lacking the positional information, thereby introducing difficulty in generation.

\textbf{Length of historical motion information.}
To determine the optimal length of historical motion information used in HMINet for controlling motion prediction, we conduct the ablation experiment in Tab.~\ref{T6}.
As shown in the table, the best results were achieved when the length was set to $n=5$.
The results gradually improve as $n<5$ and deteriorate as $n>5$.
We posit that this phenomenon arises from the inadequacy of effectively directing motion prediction when the length of historical motion information is too short, and the interference introduced by excessive information when the length of historical motion information is too long.

\begin{table}[t]
\begin{center}

\begin{tabular}{l |  c   c   c   c   c}
\toprule
Condition  & \makebox[0.05\textwidth][c]{HOTA$\uparrow$} & \makebox[0.05\textwidth][c]{IDF1$\uparrow$} & \makebox[0.05\textwidth][c]{AssA$\uparrow$} & \makebox[0.05\textwidth][c]{MOTA$\uparrow$} & \makebox[0.05\textwidth][c]{DetA$\uparrow$} \\

\midrule
$\mathbf{B}_{f-1}$     & 51.0 & \bf 49.1 & 33.3 & 88.6 &  78.5\\
$\mathbf{M}_{f-1}$     & 50.4 & 46.8 & 32.2 & 89.0 &  79.3\\
$\mathbf{I}_{f-1}$     & \bf 51.7 & 48.5 & \bf 33.8 & \bf 89.1 & \bf 79.4\\
\bottomrule
\end{tabular}
\caption{ 
Comparison of different conditions on the DanceTrack validation sets. 
The best results are shown in bold.
}
\label{T7}
\vspace{-20pt}
\end{center}
\end{table}

\begin{table}[t]
\begin{center}

\begin{tabular}{l |  c   c   c   c   c}
\toprule

$n$  & HOTA$\uparrow$ & IDF1$\uparrow$ & AssA$\uparrow$ & MOTA$\uparrow$ & DetA$\uparrow$ \\
\midrule
1      & 51.7 & 48.5 & 33.8 & 89.1 & \bf 79.4\\
2      & 52.5 & 49.9 & 34.9 & 89.2 & 79.3\\
3      & 53.0 & 52.0 & 35.9 & 89.2 & 78.6\\
5      & \bf 55.7 & \bf 55.2 & \bf 39.5 & \bf 89.3 & 78.9\\
7      & 52.5 & 49.7 & 34.9 & 89.2 & 79.3\\
10     & 51.1 & 48.7 & 33.3 & 89.1 & 78.8\\
\bottomrule
\end{tabular}
\caption{  
Evaluation of $n$ on the DanceTrack validation sets. The best results are shown in bold.
}
\label{T6}
\end{center}
\end{table}

\begin{figure}[t]
\centering
\begin{overpic}
[width=.9\linewidth,]{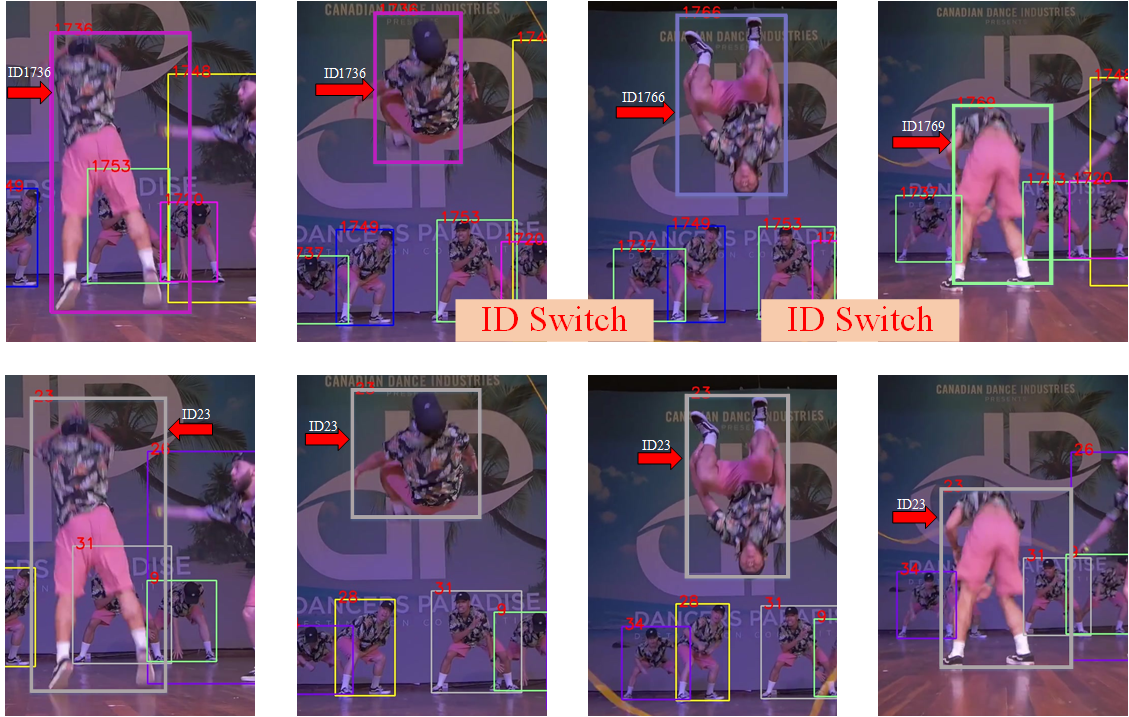}

\end{overpic}
\caption{Qualitative comparison between using KF or D$^2$MP as the motion model on the DanceTrack test set.
The upper row represents the results predicted by KF, while the lower row represents the results predicted by D$^2$MP.
The red arrow indicates the noteworthy objects.
Boxes of the same color represent the same ID.
Best viewed in color and zoom-in.
}
\vspace{-10pt}
\label{fig4}
\end{figure}

\subsection{Visualization}
Fig.~\ref{fig4} illustrates the qualitative comparison between using KF or D$^2$MP as the motion model on the test set of DanceTrack.
The upper row represents the results predicted by KF, while the lower row represents the results predicted by D$^2$MP in each case.
It is obvious that when the objects exhibit non-linear motion such as rolling, jumping, and crouching in dance, KF is unable to accurately predict the trajectories' position, resulting in the generation of new ID numbers.
In contrast, D$^2$MP exhibits greater robustness in handling these non-linear motions, accurately predicting trajectory positions, and maintaining the trajectories.
For more visualization results, please refer to the supplementary materials.

\section{Conclusion}
\label{sec5}

In this paper, we propose a diffusion-based MOT, named DiffMOT.
In contrast to previous trackers that focus on pedestrians, DiffMOT aims to track objects in non-linear motion.
To deal with the more complex non-linear pattern, we carefully design the decoupled diffusion motion predictor.
DiffMOT exceeds all previous trackers in two non-linear datasets (DanceTrack and SportsMOT) and gets comparable performances to SOTA methods on the general pedestrian-dominant dataset (MOT17). The results show that the DiffMOT has great potential in realistic applications.

\noindent\textbf{Acknowledgement.} This work was supported in part by the National Natural Science Foundation of China under Grant 62372284.

{
    \small
    \bibliographystyle{ieeenat_fullname}
    \bibliography{main}
}

\clearpage

\appendix
\setcounter{page}{11}
\maketitlesupplementary

\section{Preliminary}
The Diffusion Probabilistic Model (DPM) \cite{(34)ho2020denoising} has shown great potential in modeling non-linear mapping, yet it suffers from prolonged inference time caused by thousands of sampling steps.
DDM \cite{(38)huang2023decoupled} attempts to speed up the inference process by applying a decoupled diffusion process.
Specifically, the forward process of decoupled diffusion is split into the analytic image attenuation process and the increasing process of normal noise:
\begin{equation}
	q(\mathbf{x}_{t}|\mathbf{x}_{0}) = \mathcal{N}(\mathbf{x}_{0}+\int_{0}^{t} {\mathbf{f}_{t}\mathrm{d}t},     t\mathbf{I}),
 \label{eq1}
\end{equation} 
where $\mathbf{x}_{0}$ and $\mathbf{x}_{t}$ are the clean and noisy signals respectively, $\mathbf{f}_{t}$ denotes the analytic function representing the attenuation velocity of $\mathbf{x}_{0}$ over time $t$ ($t\in [0, 1]$), and $\mathbf{I}$ is the identity matrix. In practice, the proposed D$^2$MP uses the specific form---constant function: $\mathbf{f}_{t}=\mathbf{c}$. 
\cite{(38)huang2023decoupled} has proved that the corresponding reversed process supports sampling with arbitrary time interval $\Delta t$ and is expressed by:
\begin{equation}
	\begin{aligned}
		q(\mathbf{x}_{t-\Delta t}|\mathbf{x}_{t}, \mathbf{x}_{0}) &= \mathcal{N}(\mathbf{x}_{t} +\int_{t}^{t-\Delta t} {\mathbf{f}_{t}\mathrm{d}t}\\
		&-\frac{\Delta t}{\sqrt{t}}\boldsymbol{z}, \frac{\Delta t(t-\Delta t)}{t}\mathbf{I}),
	\end{aligned}
	\label{eq2}
\end{equation}
where $\boldsymbol{z}\sim \mathcal{N}(\mathbf{0}, \mathbf{I})$, is the noise added on $\mathbf{x}_{0}$. Actually, $\boldsymbol{z}$ and $\mathbf{f}_{t}$ are unknown in the reversed process, therefore, we need to parameterize $\mathbf{f}_{t}$ and $\boldsymbol{z}$ using a neural network $\boldsymbol{\Theta}$. In the training stage, the decoupled diffusion model uses $\boldsymbol{z}$ and $\mathbf{f}_{t}$ to supervise the parameterized $\boldsymbol{z}_{\boldsymbol{\Theta}}$ and ${\mathbf{f}}_{\boldsymbol{\Theta}}$ simultaneously:
\begin{equation}
	\min\limits_{\boldsymbol{\Theta}} \mathbb{E}_{q(\mathbf{x}_{0})} \mathbb{E}_{q(\boldsymbol{z})} [\Vert \mathbf{f}_{\boldsymbol{\Theta}}-\mathbf{f}\Vert^{2} + \Vert \boldsymbol{z}_{\boldsymbol{\Theta}}-\boldsymbol{z}\Vert^{2}].
	\label{eq3}
\end{equation}

The reversed process aims to generate $\mathbf{x}_{0}$ from $\mathbf{x}_{1}$ via Eq.~\ref{eq2} iteratively.
Due to the analyticity of the image attenuation process, we can conduct one-step sampling when $\Delta t=t=1$, removing the low speed of iterative generation.

\begin{table}[t]
\begin{center}

\begin{tabular}{l |  c   c   c   c   c}
\toprule
\makebox[0.03\textwidth][l]{Method}  & \makebox[0.04\textwidth][c]{HOTA$\uparrow$} & \makebox[0.03\textwidth][c]{IDF1$\uparrow$} & \makebox[0.03\textwidth][c]{AssA$\uparrow$} & \makebox[0.03\textwidth][c]{MOTA$\uparrow$} & \makebox[0.03\textwidth][c]{DetA$\uparrow$} \\

\midrule
\small{DiffusionTrack\cite{(39)luo2023diffusiontrack}} & 55.3 & 66.3 & 51.3 & 72.8 & 59.9\\
 MotionTrack\cite{(10)xiao2023motiontrack} & 59.7 & 71.3 & 56.8 & 76.4 & -\\
 ByteTrack\cite{(9)zhang2022bytetrack} & 61.3 & 75.2 & 59.6 &  77.8 &  63.4\\
 OC-SORT\cite{(11)cao2023observation} &  62.4 &  76.3 &  62.5 & 75.7 & 62.4\\
 StrongSORT\cite{(30)du2023strongsort}  & 62.6 & 77.0 & 64.0 & 73.8 & 61.3\\
 SparseTrack\cite{(12)liu2023sparsetrack}  & 63.5 & 77.6 & 63.1 & \bf 78.1 & \bf 64.1\\
 \small{Deep OC-SORT\cite{(14)maggiolino2023deep}}  & \bf 63.9 & \bf 79.2 & \bf 65.7 & 75.6 & 62.4\\
 DiffMOT & 61.7 & 74.9 & 60.5 & 76.7 & 63.2\\
\bottomrule
\end{tabular}
\caption{ 
Comparison with SOTA MOT trackers on the MOT20 test sets under the “private detector” protocol. 
All methods use the same YOLOX detector.
$\uparrow$ means the higher the better and $\downarrow$ means the lower the better. \textbf{Bold} numbers indicate the best result.
}
\label{T4}
\vspace{-10pt}
\end{center}
\end{table}

\section{Pseudo-code of DiffMOT}
The inference of DiffMOT consists of three parts: detection, motion prediction, and association, and the pseudo-code is shown in Alg.~\ref{alg1}.
For the $f$-th frame of the given video sequence, we use a detector to obtain the bounding boxes of objects. The detections are divided into two groups according to their confidence scores ($det.conf$). Specifically, we set two different thresholds $\tau_{high}$ and $\tau_{low}$ and group the detections by:
\begin{equation}
    \left\{
    \begin{aligned}
    \mathcal{D}_{first} = \mathcal{D}_{first} \cup \{det\} & & &det.conf \textgreater \tau_{high} \\
    \mathcal{D}_{second} = \mathcal{D}_{second} \cup \{det\} & & &\tau_{low} \textless det.conf \textless \tau_{high}.
    \end{aligned}
    \right.
\end{equation}
On the other hand, we use the proposed D$^2$MP to obtain the predicted boxes of objects in the previous trajectories.
During the association stage, we match the detected and predicted bounding boxes twice since the detections are divided into two groups.
We first match $D_{first}$ with the predicted bounding boxes via the similarity of reid features and IoU of bounding boxes. Afterwards, $D_{second}$ is matched with the predicted bounding boxes via IoU. 
The matched detections will update the trajectories. 
The unmatched tracks will be deleted. 
The unmatched detections will be initialized as new tracks. 
\begin{algorithm}[tb]
\caption{Pseudo-code of DiffMOT.}
\label{alg1}
\SetKwInOut{Input}{Input}\SetKwInOut{Output}{Output}\SetKwInOut{Parameter}{Parameter}

\Input{
A video sequence $V$;
the detector $\mathbf{D}$;
HMINet model $\mathbf{M}$; 
detection score threshold $\tau_{high}$, $\tau_{low}$;
tracking score threshold $\epsilon$}

\Parameter{Detections $\mathcal{D}_f$; predicted boxes $P_f$; pure noise $z$; objects motion $M_f$; conditions $C_f$}

\Output{Tracks $\mathcal{T}$ of the video}

Initialization: $\mathcal{T} \gets \emptyset$

\For{frame $f$ in $V$}
{
    $\mathcal{D}_{first} \gets \emptyset$; $\mathcal{D}_{second} \gets \emptyset$

    \textcolor{blue}{\tcc{Detection}}
    $\mathcal{D}_f \gets \mathbf{D}(f)$
    
    \For{$det$ in $\mathcal{D}_f$}
    {
        \If{$det.conf \textgreater \tau_{high}$}
        {$\mathcal{D}_{first} \gets \mathcal{D}_{first} \cup \{det\}$}
        \ElseIf{$\tau_{low} \textless det.conf \textless \tau_{high}$}            
        {$\mathcal{D}_{second} \gets \mathcal{D}_{second} \cup \{det\}$}
    }

    \textcolor{blue}{\tcc{Motion Prediction}}
    \For{$trk$ in $\mathcal{T}$}
    {
        $trk.C_f \gets$ motion information from $trk$

        $trk.M_f \gets \mathbf{M}(z, trk.C_f)$

        $trk.P_f \gets trk.M_f + trk.last\_location$
    }
    
    \textcolor{blue}{\tcc{Association}}
    Match $P_f$ and $D_{first}$ using reid feature and IoU

    $P_f^{remain} \gets$ remaining predicted boxes from $P_f$

    $D_f^{remain} \gets$ remaining detected boxes from $D_{first}$
    
    Match $P_f^{remain}$ and $D_{second}$ using IoU

    $P_f^{re-remain} \gets$ remaining predicted boxes from $P_f^{remain}$

    \textcolor{blue}{\tcc{Delete unmatched tracks}}
    $\mathcal{T}^{unmatched} \gets$ remaining unmatched tracks from $P_f^{re-remain}$

    $\mathcal{T} \gets \mathcal{T} \backslash \mathcal{T}^{unmatched}$

    \textcolor{blue}{\tcc{Initialize new tracks}}
    \For{$det$ in $D_f^{remain}$}
    {
        \If{$det.conf \textgreater \epsilon$}
        {$\mathcal{T} \gets \mathcal{T} \cup \{det\}$}
    }

}

Return: $\mathcal{T}$

\end{algorithm}

\section{Benchmark Evaluation on MOT20}
MOT20 \cite{(24)dendorfer2020mot20} is also one of the commonly used pedestrian-dominant datasets in MOT, characterized by higher density, and the motion is more closely approximated as linear.
We conduct the experiment on the MOT20 test sets under the "private detector" protocol to further demonstrate the performance of DiffMOT on pedestrian-dominant scenarios.
As shown in Tab.~\ref{T4}, DiffMOT achieves $61.7\%$ HOTA, $74.9\%$ IDF1, $60.5\%$ AssA, $76.7\%$ MOTA, and $63.2\%$ DetA.
The results indicate that even in pedestrian-dominant scenarios, DiffMOT, designed specifically for non-linear motion scenarios, can achieve comparable performance.

\begin{table}[t]
\begin{center}

\begin{tabular}{l l |  c   c   c}
\toprule

Train Dataset & Test Dataset & \makebox[0.05\textwidth][c]{HOTA$\uparrow$} & \makebox[0.05\textwidth][c]{IDF1$\uparrow$} & \makebox[0.05\textwidth][c]{MOTA$\uparrow$}\\
\midrule
SportsMOT & SportsMOT & 76.2 & 76.1 & 97.1\\
DanceTrack & SportsMOT & 75.6 & 75.1 & 97.1\\
\midrule
MOT17 & MOT17 & 64.5 & 79.3 & 79.8\\
DanceTrack & MOT17 & 61.8 & 74.4 & 79.1\\
\midrule
MOT20 & MOT20 & 61.7 & 74.9 & 76.7\\
DanceTrack & MOT20 & 61.2 & 73.8 & 76.2\\
\bottomrule
\end{tabular}
\caption{ 
Generalization experiments for D$^2$MP. 
}
\label{ST1}
\end{center}
\end{table}

\begin{figure}[t]
\centering
\begin{overpic}
[width=1\linewidth,]{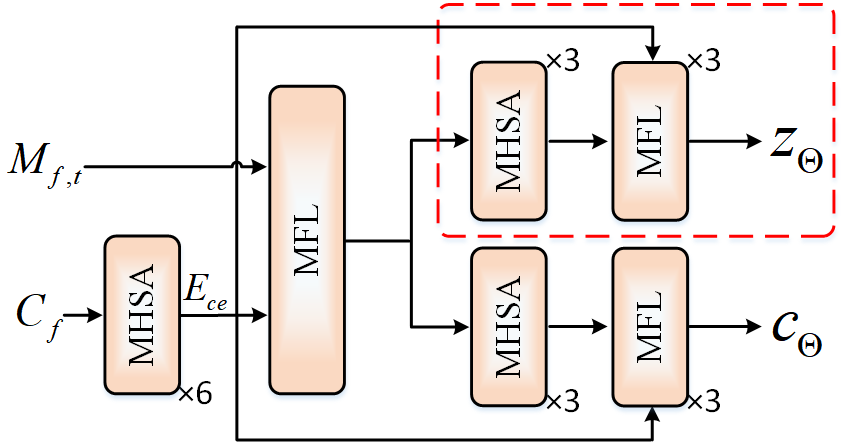}

\end{overpic}

\caption{The architecture of D$^2$MP-TB. 
The distinction from D$^2$MP-OB is enclosed within the \textcolor{red}{red} dashed box.}
\label{Sfig1}
\end{figure}

\section{Generalization of D$^2$MP}
To demonstrate the generalization of our D$^2$MP, we directly use the model trained on the DanceTrack dataset to test the performances on other datasets.
The results are shown in Tab.~\ref{ST1}.
As we can see in the table, on SportsMOT, the model trained on DanceTrack is only $0.6\%$ and $1.0\%$ lower than the model trained on SportsMOT in HOTA and IDF1.
On MOT17 and MOT20, the model trained with DanceTrack is $2.7\%$ / $0.5\%$ and $4.9\%$ / $1.1\%$ lower than the model trained on MOT17 / 20 in HOTA and IDF1.
Compared with Tab.~\ref{T2}, Tab.~\ref{T3}, and Tab.~\ref{T4}, 
the model trained solely on DanceTrack has already achieved performance comparable to the state-of-the-art methods.
Especially on SportsMOT, the model trained solely on DanceTrack has achieved SOTA performance, which outperforms previous SOTA MixSort-OC \cite{(22)Cui_2023_ICCV} by $1.4\%$ in HOTA, $0.7\%$ in IDF1, and $0.6\%$ in MOTA.
The above observation indicates that D$^2$MP possesses strong generalization capabilities, as it directly learns the distribution of all objects' motion using the diffusion model rather than learning individual object trajectories.
Our model can be applied to new scenarios without retraining, demonstrating the advantage of using the diffusion model for motion prediction.

\section{Architecture of D$^2$MP-TB}

We have conducted experiments on different architectures of D$^2$MP in the ablation study. 
The architecture of D$^2$MP-OB can be observed in Figure 3 in the manuscript.
As shown in in Fig.~\ref{Sfig1}, D$^2$MP-TB is a two-branch structure and predicts $\boldsymbol{z}_{\boldsymbol{\Theta}}$ and $\mathbf{c}_{\boldsymbol{\Theta}}$ respectively.
The distinction from D$^2$MP-OB is enclosed within the red dashed box in the figure.

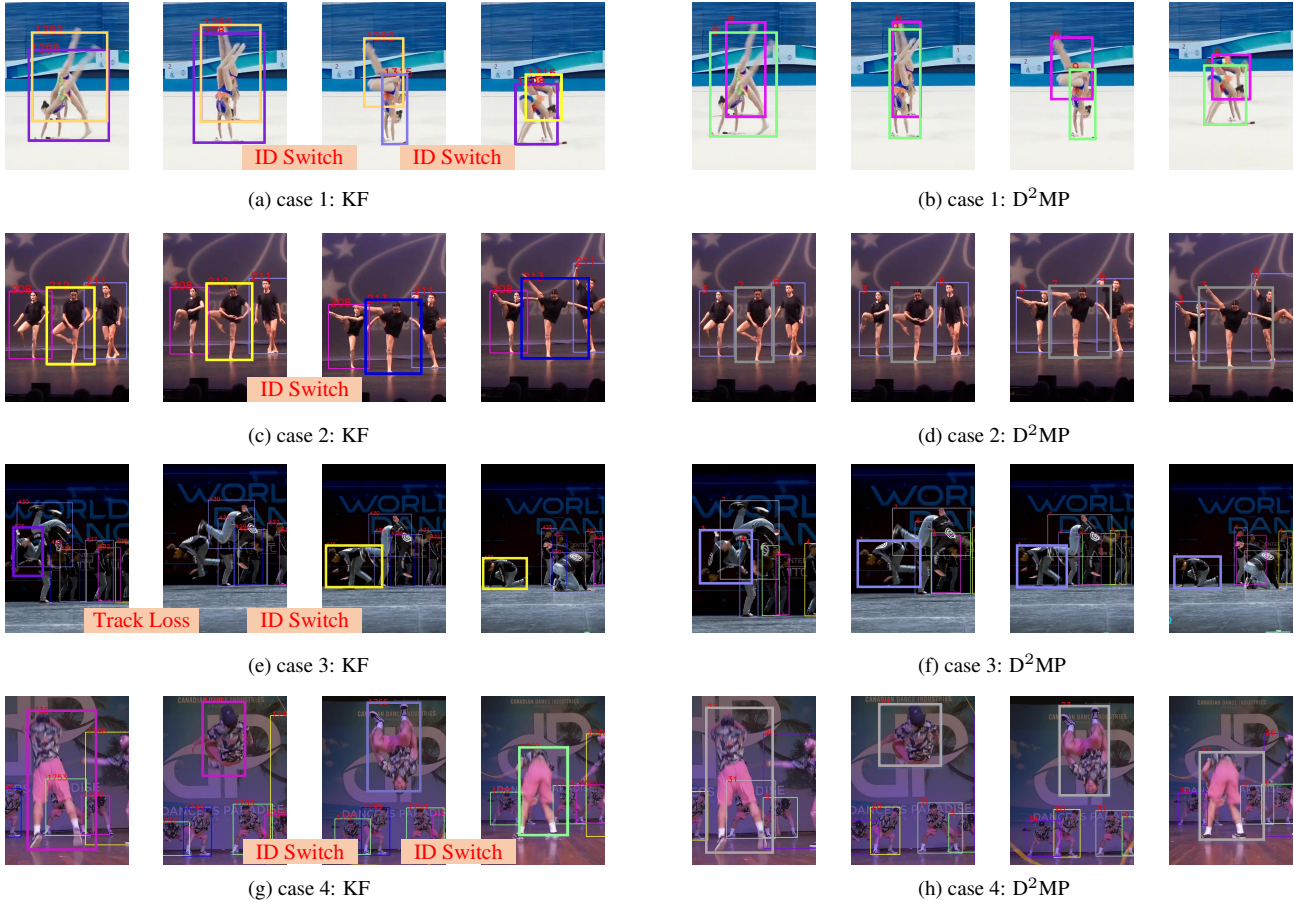
\begin{figure*}[tb]
\centering
\begin{overpic}
[width=1\linewidth,]{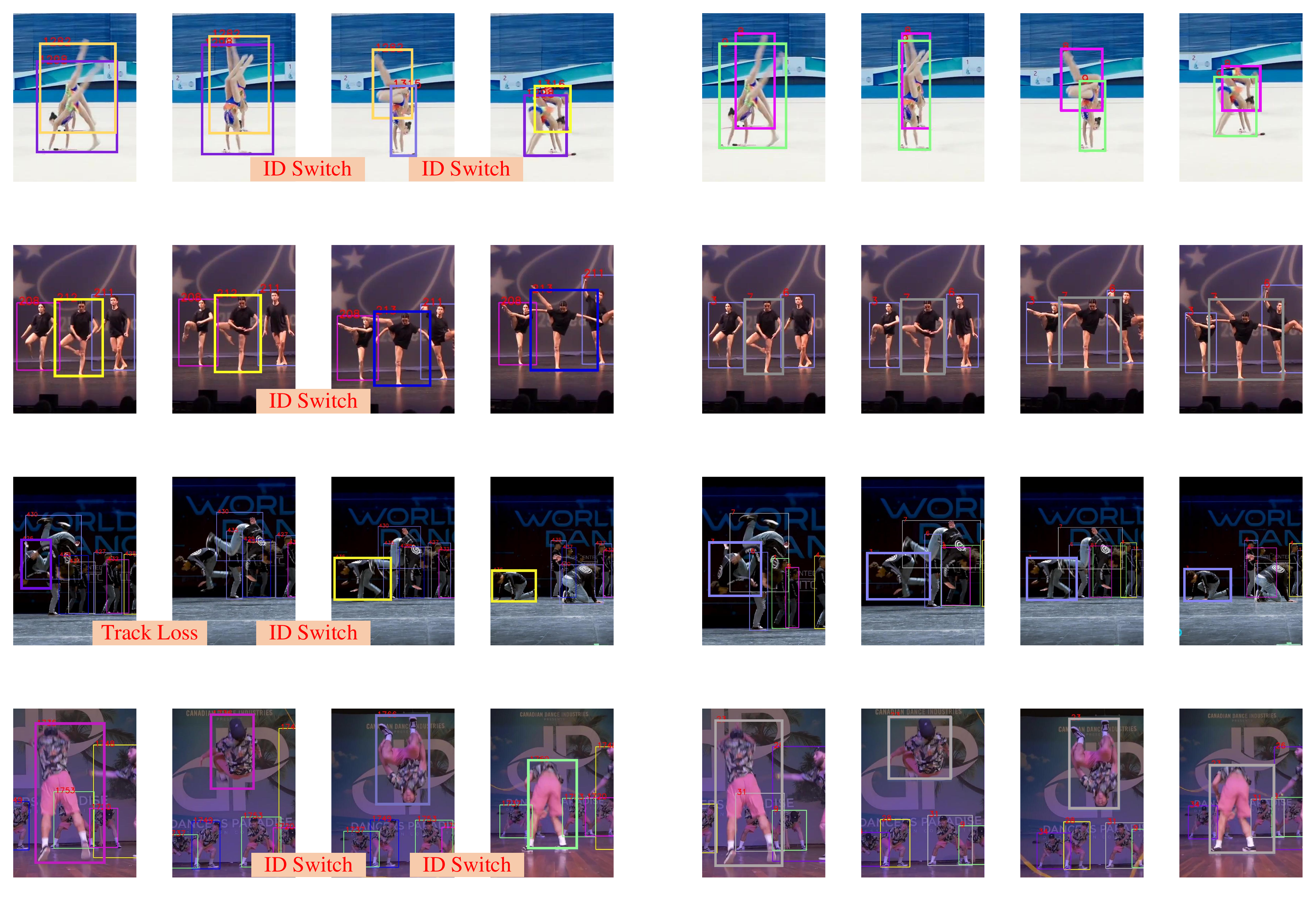}
\put(19.5,53){\footnotesize{(a) case 1: KF}}
\put(70.5,53){\footnotesize{(b) case 1: D$^2$MP}}
\put(19.5,35){\footnotesize{(c) case 2: KF}}
\put(70.5,35){\footnotesize{(d) case 2: D$^2$MP}}
\put(19.5,17.5){\footnotesize{(e) case 3: KF}}
\put(70.5,17.5){\footnotesize{(f) case 3: D$^2$MP}}
\put(19.5,0.5){\footnotesize{(g) case 4: KF}}
\put(70.5,0.5){\footnotesize{(h) case 4: D$^2$MP}}

\end{overpic}
\caption{Qualitative comparison between using KF or D$^2$MP as the motion model on the DanceTrack test set.
(a), (c), (e), and (g) represent the results using KF as the motion model.
(b), (d), (f), and (h) represent the results using D$^2$MP as the motion model.
Each pair of rows shows the comparison of the results for one sequence. 
Boxes of the same color represent the same ID.
Best viewed in color and zoom-in.
}
\vspace{-10pt}
\label{S1}
\end{figure*}

\section{More Visualization}
\subsection{Qualitative Comparison on DanceTrack}
Fig.~\ref{S1} shows some samples on the test set of DanceTrack where trackers with KF suffer from discontinuous trajectories and high ID switches while DiffMOT with D$^2$MP has strong robustness in non-linear motion scenes.
For example, the issue happens on the tracking results by trackers with KF at: (a) ID1298 $\rightarrow$ ID1315, and ID1282 $\rightarrow$ ID1316; (c) ID212 $\rightarrow$ ID213; (e) ID426 being lost and then switch to ID436; (g) ID1736 $\rightarrow$ ID1766 $\rightarrow$ ID1769.
The visual comparison indicates that when the objects exhibit non-linear motions in dance scenarios, trackers with KF are unable to predict the accurate trajectories' position, resulting in a large ID switch.
In contrast, D$^2$MP exhibits greater robustness in handling these non-linear motions.

\subsection{Qualitative Comparison on SportsMOT}
\begin{figure*}[tb]
\centering
\begin{overpic}
[width=1\linewidth,]{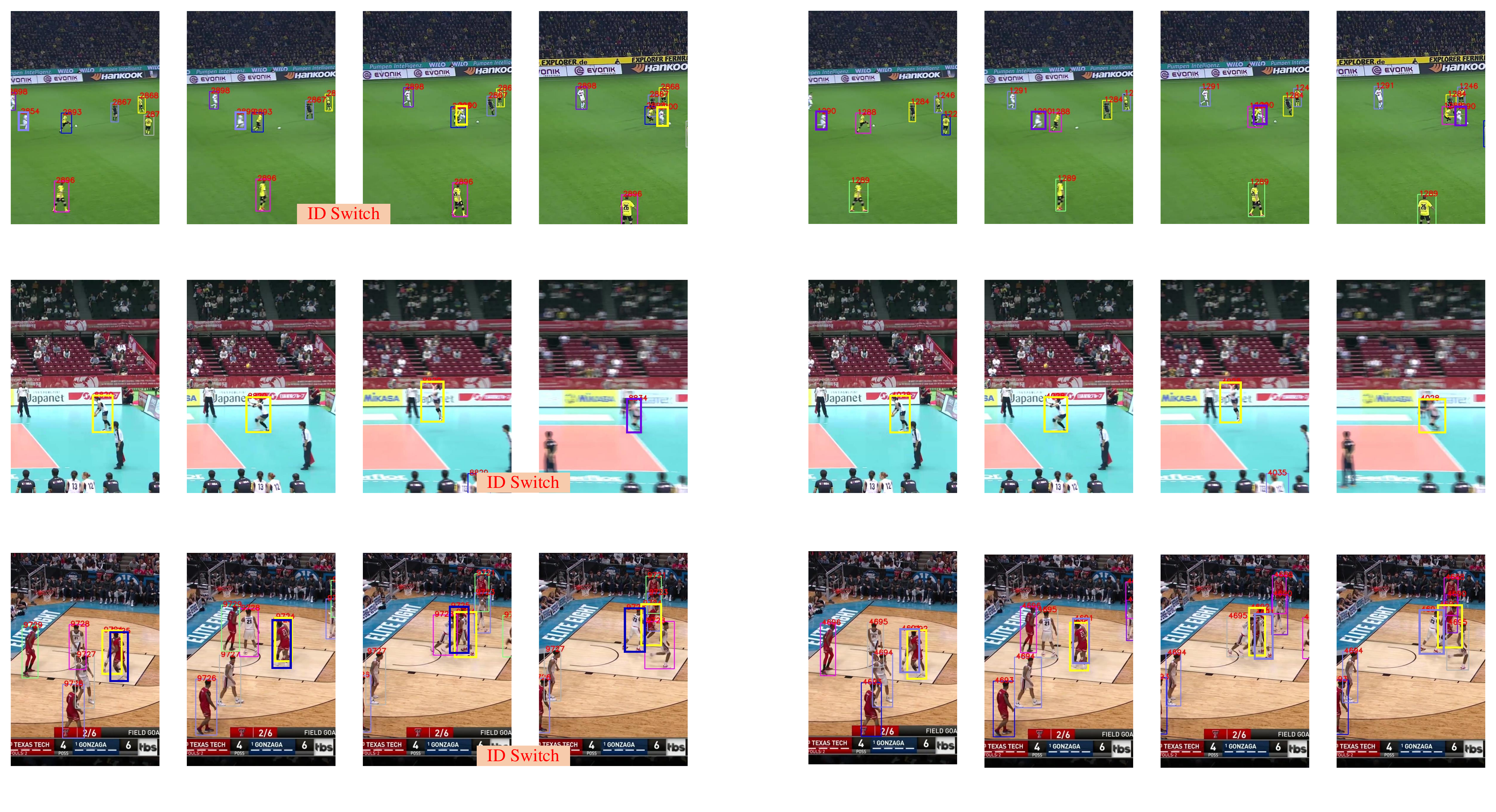}

\put(16,36.5){\footnotesize{(a) Football case: KF}}
\put(66,36.5){\footnotesize{(b) Football case: D$^2$MP}}
\put(16,18.5){\footnotesize{(c) Volleyball case: KF}}
\put(66,18.5){\footnotesize{(d) Volleyball case: D$^2$MP}}
\put(16,0.5){\footnotesize{(e) Basketball case: KF}}
\put(66,0.5){\footnotesize{(f) Basketball case: D$^2$MP}}

\end{overpic}
\vspace{-20pt}
\caption{Qualitative comparison between using KF or D$^2$MP as the motion model on the SportsMOT test set.
(a), (c), and (e) represent the results using KF as the motion model.
(b), (d), and (f) represent the results using D$^2$MP as the motion model.
Each pair of rows shows the comparison of the results for one sequence. 
All of the cases are in scenarios with a moving camera.
Boxes of the same color represent the same ID.
Best viewed in color and zoom-in.
}
\label{S2}
\end{figure*}

Fig.~\ref{S2} depicts more qualitative comparisons between employing KF and D$^2$MP as the motion model on the test set of SportsMOT.
We select samples from diverse scenes, including football, volleyball, and basketball scenes.
The issue happens on the tracking results by trackers with KF at: (a) ID2854 $\rightarrow$ ID2900; (c) ID8820 $\rightarrow$ ID8834; (e) ID switch between ID9724 and ID9725.
The visual comparison in the figure highlights that when the objects exhibit non-linear motions such as acceleration or deceleration in sports scenarios, KF often hard to provide accurate predictions, while DiffMOT demonstrates the ability to predict the objects' position accurately in such scenarios.

\subsection{Visual Results on MOT17/20}
\begin{figure*}[tb]
\centering
\begin{overpic}
[width=1\linewidth,]{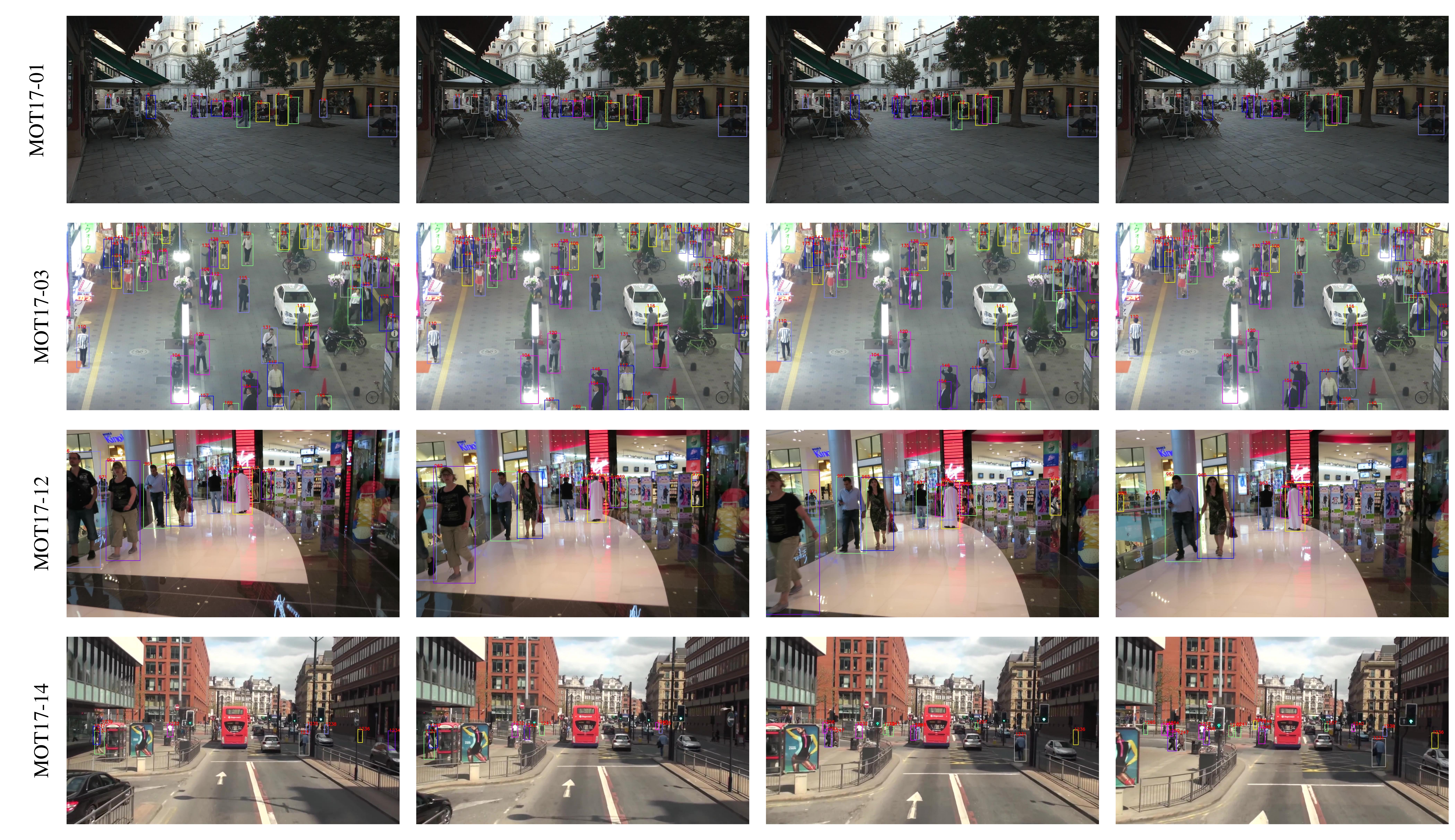}

\end{overpic}
\caption{The visualization of DiffMOT tracking results on the test set of MOT17.
Boxes of the same color represent the same ID.
Best viewed in color and zoom-in.
}
\label{S3}
\end{figure*}

\begin{figure*}[tb]
\centering
\begin{overpic}
[width=1\linewidth,]{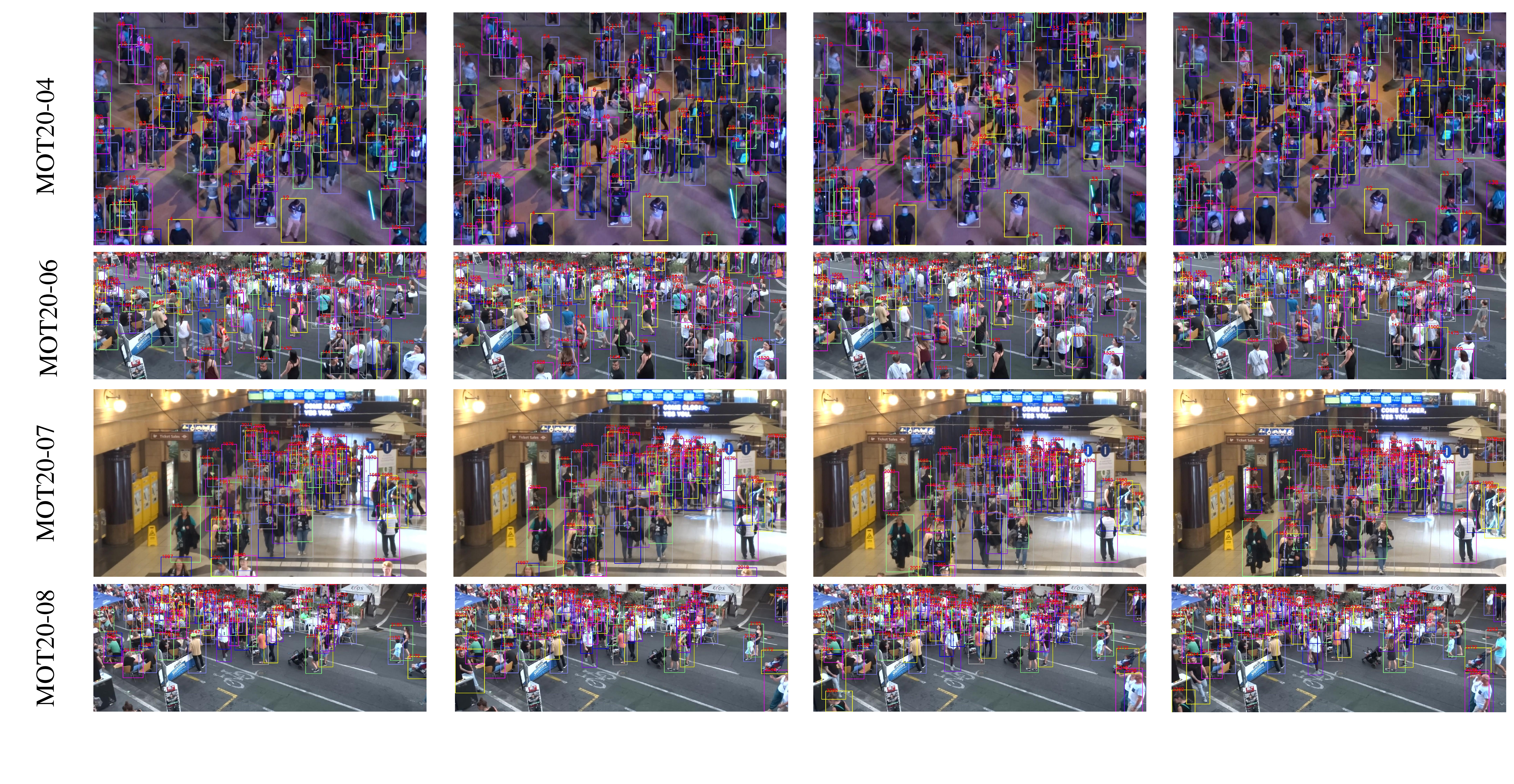}

\end{overpic}
\caption{The visualization of DiffMOT tracking results on the test set of MOT20.
Boxes of the same color represent the same ID.
Best viewed in color and zoom-in.
}
\label{S4}
\end{figure*}

Fig.~\ref{S3} and Fig.~\ref{S4} show several tracking results of our DiffMOT on the test set of MOT17 and MOT20, respectively.
It can be observed that Although the proposed DiffMOT is designed specifically for non-linear motion scenes, it can still achieve appealing results.

\section{Illustration of Failure Cases}
\begin{figure*}[tb]
\centering
\begin{overpic}
[width=1\linewidth,]{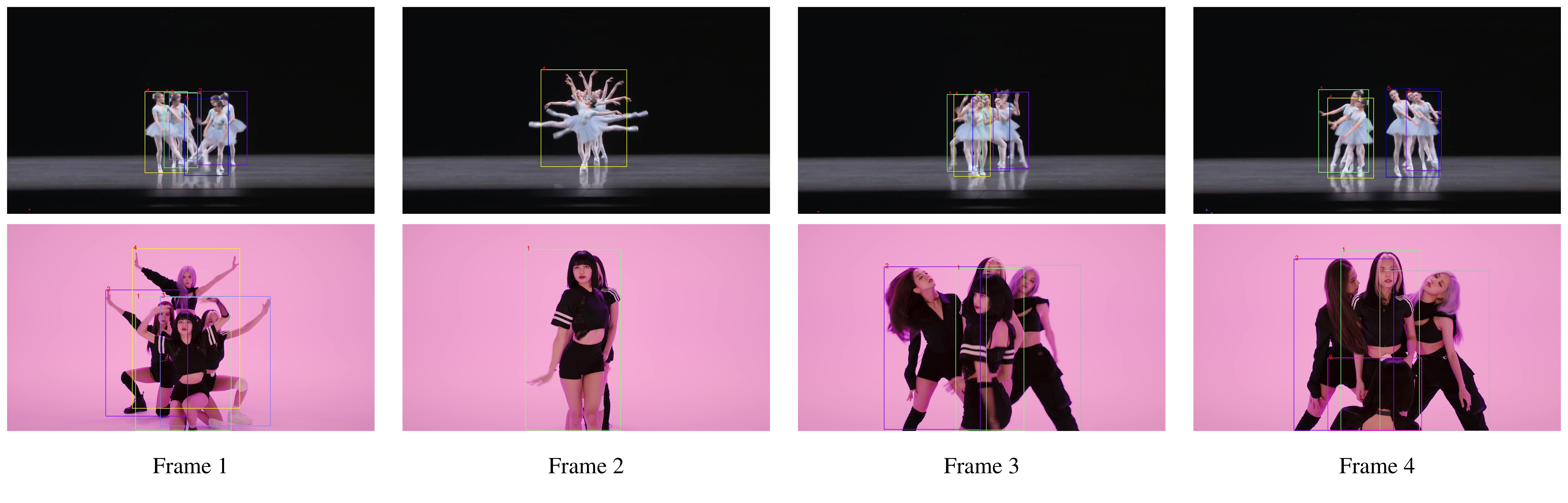}

\end{overpic}
\caption{Illustration of two failure cases. We show the two most common failure cases of our approach.
In the first row, due to the absence of velocity direction constraints, ID switches have occurred.
In the second row, due to the difficulty in recovering long-term lost objects, the new ID is generated.
}
\label{S5}
\end{figure*}

We visualize two failure cases in Fig.~\ref{S5}.
For the first case, when different objects are passing through each other, objects with ID "1" and “2”, as well as objects with ID “4“ and “5”, underwent an exchange of identities.
This is due to the absence of velocity direction constraints in the motion model.
We believe that incorporating velocity direction constraints to restrict the generation of predicted boxes could help address this issue.

In the second case, the object with ID "3" disappears in "Frame 2" and reappears in "Frame 3" as ID "7". Simultaneously, the object with ID "4" disappears in "Frame 2" and reappears in "Frame 4" as ID "1", while the object with original ID "1" becomes a new ID "8".
This phenomenon occurs due to the difficulty in recovering long-term lost objects of our motion model.
When an object is lost for an extended period, it becomes challenging to re-associate the object accurately, leading to the generation of new IDs or ID switches.
In our future work, we intend to explore the generation of multi-frame trajectories to improve the motion model's capacity for long-term matching.

\end{document}